\documentclass[11pt]{article}
\usepackage{multicol}
\usepackage{array}
\usepackage[table]{xcolor}
\usepackage[most]{tcolorbox}
\usepackage[final]{acl}
\usepackage{multirow}
\usepackage{hyperref}
\usepackage{wrapfig}
\usepackage{times}
\usepackage{latexsym}
\usepackage[ruled,vlined]{algorithm2e}
\usepackage{enumitem}
\usepackage{url}
\usepackage[T1]{fontenc}
\usepackage{booktabs}
\usepackage{pifont}
\usepackage{soul}
\definecolor{errorboxbg}{RGB}{255,245,230}
\definecolor{errorboxborder}{RGB}{255,120,60}
\definecolor{errorboxheader}{RGB}{255,200,150}
\usepackage{algpseudocode}
\usepackage{tabularx} 
\usepackage{ragged2e}

\newtcolorbox{errorstudybox}[1][]{
  colback=errorboxbg,
  colframe=errorboxborder,
  coltitle=black,
  colbacktitle=errorboxheader,
  fonttitle=\bfseries,
  title=Error Study,
  sharp corners,
  boxrule=1pt,
  left=2mm, right=2mm, top=1mm, bottom=1mm,
  enhanced,
  attach boxed title to top left={yshift=-2mm, xshift=2mm},
  boxed title style={size=small, colframe=errorboxborder, colback=errorboxheader},
  #1
}

\usepackage[utf8]{inputenc}

\usepackage{microtype}

\usepackage{inconsolata}
\usepackage{amsmath}
\usepackage{amssymb}
\usepackage{graphicx}
\usepackage[most]{tcolorbox}
\usepackage{tikz}      
\usepackage{fontawesome}
\usetikzlibrary{shadows}  
\usetikzlibrary{positioning}
\tcbset{
    case/.style={
    colback=gray!5!white, 
        colframe=white, 
        boxrule=0.5mm, 
        left=1mm, 
        right=1mm, 
        top=1mm, 
        bottom=1mm, 
        arc=3mm, 
        boxsep=3mm,
        drop shadow={black!50!white}, 
        enhanced,
        overlay={
            \node[fill=cyan!70!black, text=white, rounded corners=1mm, font=\bfseries\scriptsize, inner sep=1mm] 
            at (frame.north west) 
            [xshift=15mm, yshift=-0mm] {Level-1 Example};
            }
    }
}

\tcbset{
    details/.style={
        colback=gray!5!white, 
        colframe=white, 
        boxrule=0.5mm, 
        left=1mm, 
        right=1mm, 
        top=1mm, 
        bottom=1mm, 
        arc=3mm, 
        boxsep=3mm,
        drop shadow={black!50!white}, 
        enhanced,
        overlay={
            \node[fill=purple!70!white, text=white, rounded corners=1mm, font=\bfseries\scriptsize, inner sep=1mm] 
            at (frame.north west) 
            [xshift=15mm, yshift=-0mm] {Level-2 Example};
        }
    }
}

\tcbset{
    level3/.style={
        colback=gray!5!white, 
        colframe=white, 
        boxrule=0.5mm, 
        left=1mm, 
        right=1mm, 
        top=1mm, 
        bottom=1mm, 
        arc=3mm, 
        boxsep=3mm,
        drop shadow={black!50!white}, 
        enhanced,
        overlay={
            \node[fill=orange!70!white, text=white, rounded corners=1mm, font=\bfseries\scriptsize, inner sep=1mm] 
            at (frame.north west) 
            [xshift=15mm, yshift=-0mm] {Level-3 Example};
        }
    }
}

\tcbset{
    prompt/.style={
        colback=blue!5!white,         
        colframe=blue!75!black,       
        boxsep=4pt,                   
        arc=5pt,                      
        boxrule=0.8pt,                
        width=\linewidth,
        enhanced,
        drop shadow={blue!30!black},  
        boxed title style={
            colback=blue!20!white,    
            colframe=blue!50!black,   
            boxrule=0.8pt,            
            rounded corners=south,    
            size=small,
            boxsep=3pt,               
            left=3mm,
            right=3mm,
            fontupper=\color{blue!30!black}\bfseries  
        },
        attach boxed title to top center={yshift=-\tcboxedtitleheight/2},  
        top=3mm,
        bottom=3mm,
        coltitle=blue!40!black        
    }
}

\title{\textit{PolicyLLM}: Towards Excellent Comprehension of Public Policy for Large Language Models}

\author{
 \textbf{Han Bao\textsuperscript{1}},
 \textbf{Penghao Zhang\textsuperscript{2}},
 \textbf{Yue Huang\textsuperscript{1}},
 \textbf{Zhengqing Yuan\textsuperscript{1}},
\\
 \textbf{Yanchi Ru\textsuperscript{1}},
 \textbf{Rui Su\textsuperscript{1}},
 \textbf{Yujun Zhou\textsuperscript{1}},
 \textbf{Xiangqi Wang \textsuperscript{1}},
\\
 \textbf{Kehan Guo\textsuperscript{1}},
 \textbf{Nitesh V Chawla\textsuperscript{1}},
 \textbf{Yanfang Ye\textsuperscript{1}},
 \textbf{Xiangliang Zhang\textsuperscript{1}},
\\
 \textsuperscript{1}University of Notre Dame,
 \textsuperscript{2}Independent Researcher
\\
 \small{
   \textbf{Correspondence:} \href{mailto:hbao@nd.edu}{hbao@nd.edu}
 }
}

\begin{document}
\maketitle

\begin{abstract}
Large Language Models (LLMs) are increasingly integrated into real-world decision-making, including in the domain of public policy. Yet, their ability to comprehend and reason about policy-related content remains underexplored. To fill this gap, we present \textbf{\textit{PolicyBench}}, the first large-scale cross-system benchmark (US-China) evaluating policy comprehension, comprising 21K cases across a broad spectrum of policy areas, capturing the diversity and complexity of real-world governance. Following Bloom's taxonomy, the benchmark assesses three core capabilities: (1) \textbf{Memorization}: factual recall of policy knowledge, (2) \textbf{Understanding}: conceptual and contextual reasoning, and (3) \textbf{Application}: problem-solving in real-life policy scenarios. Building on this benchmark, we further propose \textbf{\textit{PolicyMoE}}, a domain-specialized Mixture-of-Experts (MoE) model with expert modules aligned to each cognitive level. The proposed models demonstrate stronger performance on application-oriented policy tasks than on memorization or conceptual understanding, and yields the highest accuracy on structured reasoning tasks. Our results reveal key limitations of current LLMs in policy understanding and suggest paths toward more reliable, policy-focused models\footnote{Accepted as a Findings paper at ACL 2026. Dataset has been released at \url{https://github.com/wad3birch/PolicyLLM}.}.

\end{abstract}

\section{Introduction}

In recent years, Large Language Models (LLMs) \cite{vaswani2017attention,touvron2023llama,achiam2023gpt} have achieved remarkable progress, demonstrating intelligent and superior performance in a wide range of natural language processing (NLP) tasks, including machine translation \cite{zhu2024multilingual}, code generation \cite{svyatkovskiy2020intellicode,chen2021evaluating}, and article writing \cite{yuan2022wordcraft}. In parallel with these advancements, a growing body of research has focused on systematically benchmarking LLM capabilities across multiple cognitive dimensions, including language understanding \cite{wangglue,wang2024mmlu}, reasoning \cite{xiang2023beyond,cobbe2021training,joshi2017triviaqa}, and knowledge acquisition \cite{yang2015wikiqa}.

Beyond traditional NLP tasks, LLMs are increasingly being deployed in high-stakes real-world decision-making contexts, such as education \cite{xiao2023evaluating}, law \cite{fei2024lawbench, guha2023legalbench, zhou2024lawgpt}, healthcare \cite{tang2023does} and public administration \cite{pesch2025potentials}. Among these, public policy stands out as particularly consequential: supporting policy analysis and generation requires not only factual knowledge, but also contextual reasoning and value-sensitive judgment \citep{hou2025urban}. Missteps can have tangible social consequences—for example, a model that miscalculates rural funding allocations by relying on the wrong fiscal base may cause substantial under-allocation of resources. Ensuring that LLMs develop a reliable and nuanced understanding of policy content is therefore both a technical necessity and an ethical imperative.

Understanding and applying public policy presents unique challenges for LLMs. While the field’s interdisciplinary nature, contextual dependence, and linguistic complexity are well recognized, the central obstacles to advancing policy-aware AI can be framed as a three-tiered problem.
1) \textit{\textbf{The Evaluation Challenge: Lack of Rigorous Benchmarks.}}
There is currently no comprehensive benchmark to systematically assess the policy comprehension capabilities of LLMs. Without standardized evaluation frameworks, it is difficult to measure performance across skills ranging from factual recall to conceptual reasoning and practical application, hindering objective comparison and targeted improvement.
2) \textit{\textbf{The Diagnostic Challenge: Identifying Strengths and Weaknesses.}}
Aggregate metrics obscure where models succeed and fail. It remains unclear which cognitive abilities, policy domains, or linguistic contexts pose the greatest difficulties. Fine-grained diagnostic analysis is therefore essential to pinpoint strengths and weaknesses.
3) \textit{\textbf{The Adaptation Challenge: Developing Specialized Models.}}
General-purpose LLMs often struggle with the distinct demands of policy tasks. A key challenge is how to adapt existing architectures to better handle the multifaceted requirements of policy analysis, thereby closing the gaps revealed through rigorous evaluation and diagnosis.

To rigorously evaluate the gap, we present \textit{PolicyBench}, a cross-system benchmark (US-China) specifically designed to assess LLMs' understanding of public policy in both China and the United States. \textit{PolicyBench} encompasses a broad spectrum of policy domains and features meticulously crafted questions targeting three cognitive levels: memorization, understanding, and application. Through extensive experiments, we find that model performance improves steadily from memorization to application tasks, with LLMs showing particular strength in structured reasoning scenarios such as numerical calculation and scenario-based decision-making, while still facing challenges in abstract or ambiguous policy contexts and in handling Chinese policy texts. To further enhance LLMs’ policy-related reasoning, we propose \textit{PolicyMoE}---a MoE model~\cite{jacobs1991adaptive,jordan1994hierarchical} trained on policy-focused data~\cite{kang2024self}. \textit{PolicyMoE} integrates three specialized expert models, each excelling in distinct capabilities. Experimental results demonstrate that \textit{PolicyMoE} significantly outperforms general-purpose LLMs on policy tasks.

Overall, our main contributions are as follows:

\begin{itemize}[noitemsep,leftmargin=0pt]
    \item[$\triangleright$] We construct \textit{PolicyBench}, a comprehensive cross-system benchmark for evaluating LLMs' policy understanding across diverse domains---in both Chinese and US contexts.
    \item[$\triangleright$] Through extensive experiments and human evaluation on \textit{PolicyBench}, we uncover key findings on the strengths and limitations of LLMs in cross-system policy understanding.
    \item[$\triangleright$] We propose \textit{PolicyMoE}, an MoE model fine-tuned on \textit{PolicyBench}, which achieves superior performance over strong baselines and underscores the potential of domain-adaptive pretraining for governance-related tasks.
\end{itemize}

\begin{figure*}[t]
    \centering
  \includegraphics[width=\linewidth]{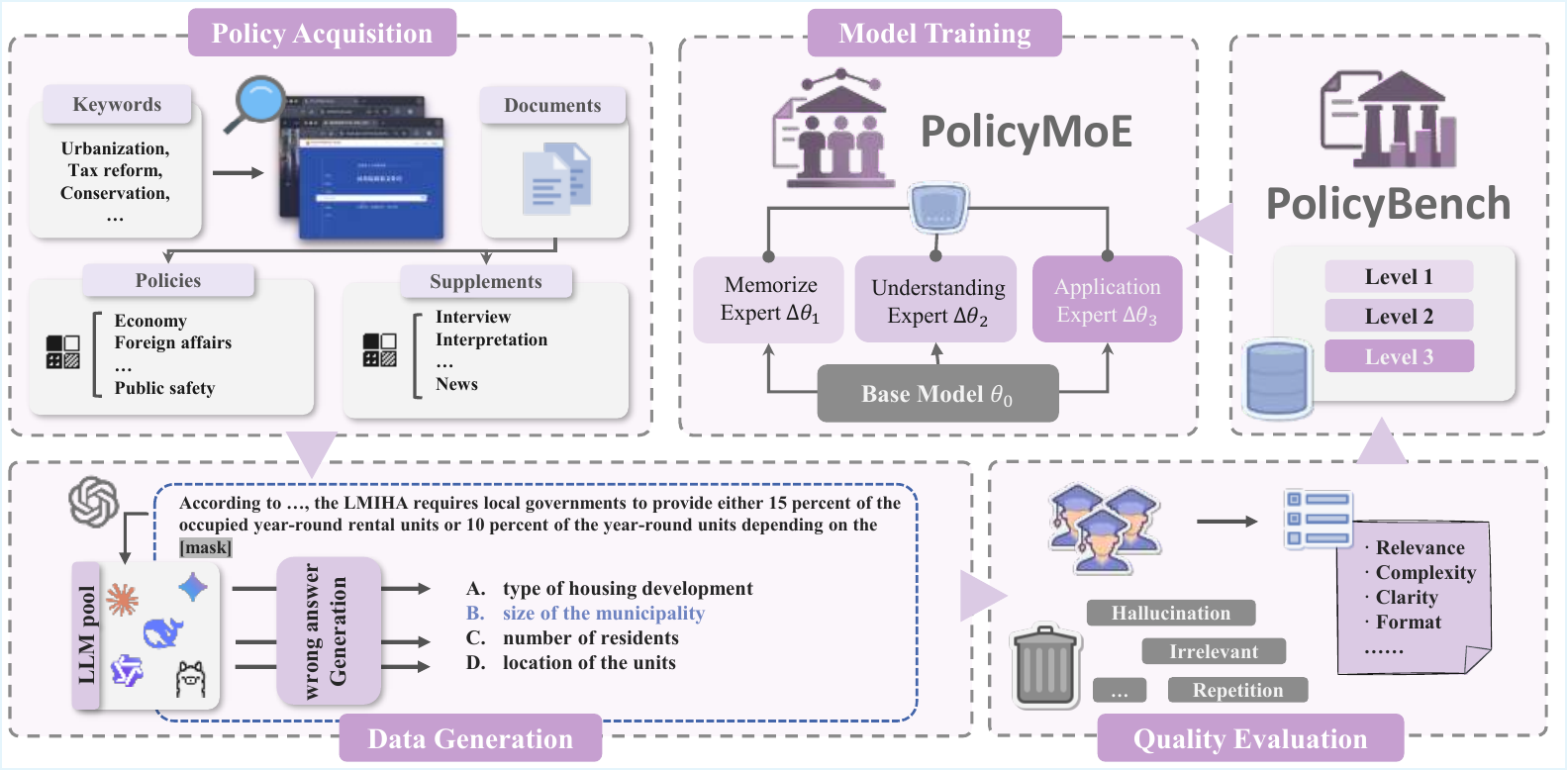}
  \caption{Three levels of evaluating LLM in \textit{PolicyBench}.}
  \label{fig:policybench}
  \vspace{-15pt}
\end{figure*}

\vspace{-8pt}
\section{The \textit{PolicyBench}}

In this section, we provide a detailed introduction to the design and construction principles of \textit{PolicyBench}. To construct our benchmark, \textbf{we focused on two of the world's most significant yet distinct policy environments: mainland China (CN) and the United States federal government (US)}. This deliberate selection provides a high-contrast, cross-system testbed for evaluating an LLM's core policy comprehension capabilities across different governance systems. While we acknowledge this does not encompass the full global policy frameworks, it establishes a critical and challenging baseline for this foundational area.

\subsection{Policy Acquisition}
In the process of policy collection, we initially gathered a broad set of Chinese and US policy documents and related materials. To ensure relevance and timeliness of the content, we applied a filtering process that removed outdated policies, duplicate entries, and documents not related to substantive policy content (\textit{ e.g., purely procedural notices or administrative logistics}), details in \autoref{sec:data}. After this curation step, we retained \textbf{721} Chinese policies and \textbf{1,890} supplementary Chinese policy materials (\textit{e.g., official commentaries, media news, expert interviews, and public consultations}), as well as \textbf{603} US policies and \textbf{1,082} supplementary US materials:

\begin{itemize}
    \item[$\bullet$] For Chinese policies, all documents were sourced exclusively from the Policy Document Repository of the State Council of China.\footnote{\url{https://www.gov.cn/zhengce/zhengcewenjianku/}} To categorize the policies, we first followed the organizational structure of the State Council, then refined it to better reflect the content distribution within the corpus. Based on this adapted structure, we grouped the policies into eight domains (\emph{e.g., Public safety}). To retrieve relevant materials, we selected representative search terms—such as “Belt and Road Initiative” and “Double Reduction Policy”—based on “Hot Words” highlighted by major official media and platforms, see \autoref{fig:keywords} for details. In addition, we collected supplementary materials including official interpretations, policy outcomes, and expert interviews from extended social media sources. 
    \item[$\bullet$] For US policies: As there is no centralized repository for federal policies in the US, we collected policy documents from the official websites of 12 US federal departments (Details in \autoref{tab:data_sources}). Supplementary materials were gathered from authoritative news outlets such as \textit{Reuters}, \textit{Fox News}, etc. 
\end{itemize}

All collected policies fall within the timeframe of 2000 to January 2025, with a primary focus on the most recent decade. All operations fully complied with ethical standards, and all data were lawfully sourced from open-access public databases.

\begin{figure*}[h]
    \centering
    \includegraphics[width=1.0\linewidth]{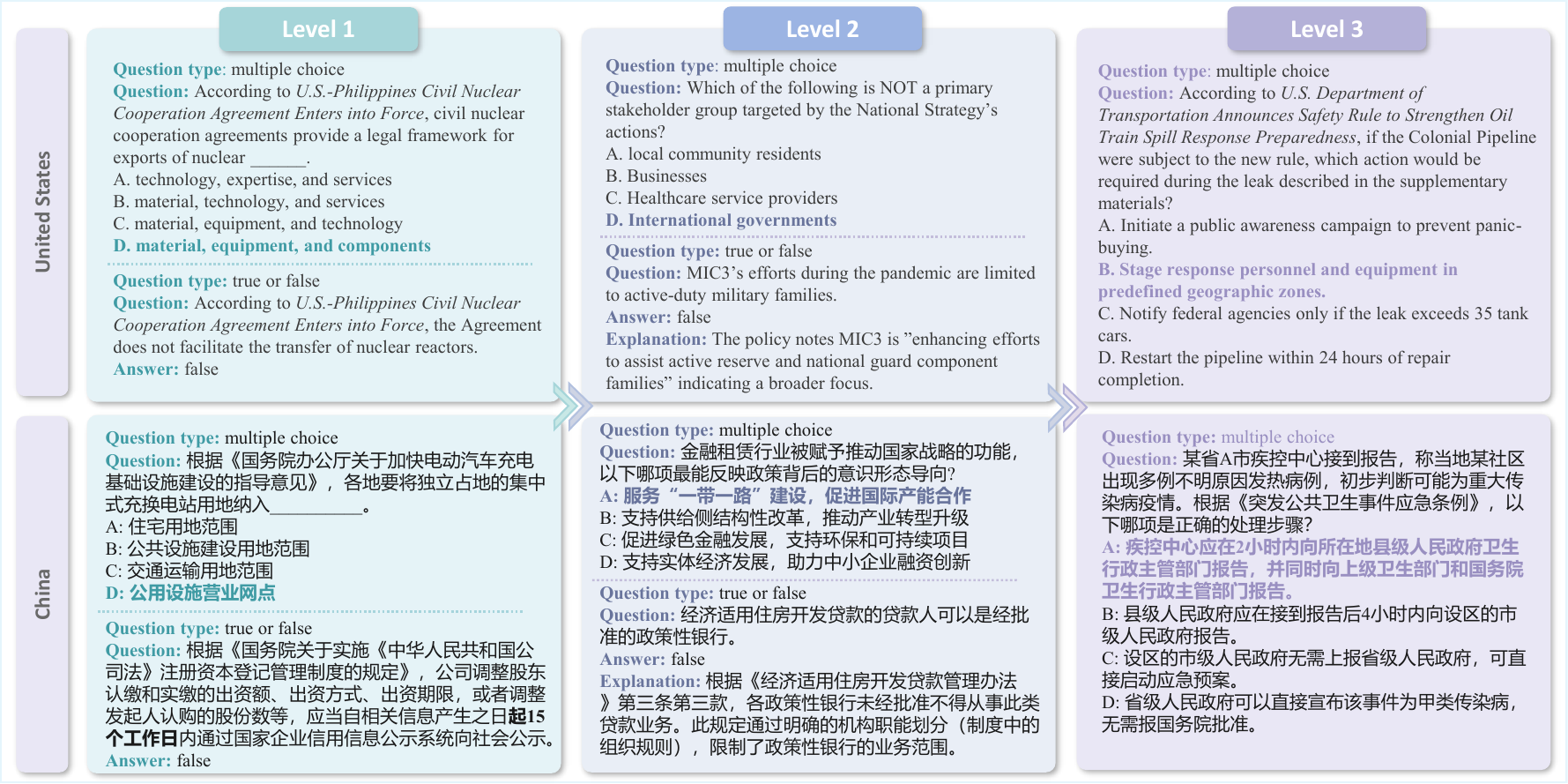}
    \caption{Selected examples from PolicyBench spanning three levels and two languages.}
    \label{fig:case-study}
\end{figure*}






\begin{table}[htbp]
\centering
\scriptsize 
\renewcommand{\arraystretch}{1.1}
\setlength{\tabcolsep}{2pt} 
\caption{Task list of \textit{PolicyBench} and respective numbers (``Mem'' = Memorization, ``Und'' = Understanding, ``App'' = Application).}
\label{tab:fined_grained_task}
\begin{tabular}{
  >{\centering\arraybackslash}p{0.6cm} 
  >{\centering\arraybackslash}p{0.6cm} 
  >{\centering\arraybackslash}p{4.0cm} 
  >{\centering\arraybackslash}p{0.8cm} 
  >{\centering\arraybackslash}p{0.8cm}
}
\toprule
\textbf{Level} & \textbf{ID} & \textbf{Task} & \multicolumn{2}{c}{\textbf{Number}} \\
\cmidrule(lr){4-5}
& & & \textbf{CN} & \textbf{US} \\
\midrule

\multirow{3}{*}{\textbf{Mem.}} 
& 1-1 & \scriptsize Article/Date Memorization & 4,498 & 3,351 \\
& 1-2 & \scriptsize Terminology Recognition & 237 & 126 \\
& 1-3 & \scriptsize Organization Identification & 268 & 228 \\

\midrule
\multirow{3}{*}{\textbf{Und.}} 
& 2-1 & \scriptsize Idea Understanding & 1,219 & 1,013 \\
& 2-2 & \scriptsize Interest Understanding & 1,145 & 996 \\
& 2-3 & \scriptsize Institution Understanding & 1,620 & 1,484 \\

\midrule
\multirow{4}{*}{\textbf{App.}} 
& 3-1 & \scriptsize Policy-Based Numerical Reasoning & 918 & 452 \\
& 3-2 & \scriptsize Scenario-Based Decision-Making & 77 & 179 \\
& 3-3 & \scriptsize Procedural/Institutional Implementation & 320 & 391 \\
& 3-4 & \scriptsize Policy Logic and Value Explanation & 1,199 & 1,214 \\

\bottomrule
\end{tabular}
\end{table}

\subsection{Dataset Curation}

To facilitate fine-grained analysis of model capabilities, we categorize the benchmark into \textbf{10 task types}, each targeting a distinct subskill relevant to public policy comprehension. This taxonomy enables a comprehensive assessment of LLMs across a wide range of cognitive and policy domains.

The categorization is structured around a three-level hierarchy informed by Bloom’s Taxonomy of Educational Objectives~\citep{krathwohl2002revision}, a widely recognized framework in cognitive and educational psychology. Bloom’s taxonomy organizes cognitive skills from basic factual recall to deeper understanding and the application of knowledge in real-world contexts. Drawing on this structure, our benchmark defines three assessment levels:

\begin{itemize}[noitemsep,leftmargin=0pt]
    \item \textbf{Level 1: Memorization.} This level focuses on factual recall, such as memorizing publication dates, institutional actors, specific provisions, or technical terminology. Tasks at this level require minimal inference and primarily test a model’s ability to retrieve explicit information from policy documents.
    \item \textbf{Level 2: Understanding.} At this tier, we move beyond literal recall to examine a model’s capacity for conceptual understanding and contextualization. Guided by the \textbf{3I framework} from policy studies, which emphasizes Ideas, Interests, and Institutions~\citep{hall1996political}. Level 2 tasks probe how well a model can interpret underlying motivations, identify key stakeholders, and comprehend institutional logic within policies.
    \item \textbf{Level 3: Application.} The highest level assesses the model’s ability to apply policy knowledge in practical scenarios. Tasks at this tier require reasoning about hypothetical or real-world situations, evaluating implications, and suggesting appropriate actions based on policy content.
\end{itemize}


As summarized in \autoref{tab:fined_grained_task}, this hierarchical task design enables a fine-grained evaluation of policy comprehension across multiple cognitive levels. Annotation guidelines and quality control procedures detailed in \autoref{sec:human_eval}, with formal task definitions are provided in \autoref{app:task_definition}.

\vspace{-5pt}
\subsection{Detail of Distractor Generation}

To mitigate the bias that can arise when a single LLM generates all distractors, we employ a heterogeneous \emph{model pool} and harvest incorrect options in an iterative fashion. With details mathmatically referred in Algorithm~\ref{alg:distractor_generation} and given a stem--answer pair $\langle q,\;a_{\text{gold}}\rangle$, we initially sample a model $m$ from the pool and prompt it with the original question while \emph{explicitly labelling} $a_{\text{gold}}$ as a prior incorrect answer; the model is instructed to propose a new, plausible response that must differ from $a_{\text{gold}}$. If the resulting candidate $d$ is neither redundant nor equal to the gold answer, it is added to the distractor set $D$. We iteratively resample additional models and repeat the procedure, supplying both the original question and the accumulated distractors, until the total number of distractors satisfies $\lvert D\rvert = k - 1$. Finally, we randomly permute $a_{\text{gold}}\cup D$ to form the list of $k$ options.

\begin{algorithm}[t]
\caption{Distractor Generation for Multiple-Choice Question Construction}
\label{alg:distractor_generation}
\KwIn{Question $q$, Correct Answer $a_{\text{gold}}$, LLM pool $\mathcal{M}$, Target number of choices $k = 4$}
\KwOut{Multiple-choice question with one correct answer and $k-1$ distractors}
Initialize distractor set $\mathcal{D} \gets \emptyset$\;

\While{$|\mathcal{D}| < k - 1$}{
    Sample a model $m \sim \mathcal{M}$\;
    
    Construct prompt:\;
    \Indp
        Include the question $q$\;
        Provide $a_{\text{gold}}$ as a previous incorrect answer\;
        Instruct model $m$ to generate a new plausible answer that is \textbf{also incorrect}\;
    \Indm
    
    Generate distractor candidate $d \gets m(q, a_{\text{gold}} \text{ marked as incorrect})$\;
    
    \If{$d \notin \mathcal{D}$ and $d \neq a_{\text{gold}}$}{
        Add $d$ to $\mathcal{D}$\;
    }
}
Randomly shuffle $a_{\text{gold}} \cup \mathcal{D}$ to form final options\;
\Return $\{q, \text{options}, a_{\text{gold}}\}$\;
\end{algorithm}

\section{The \textit{PolicyMoE}}

Specifically in policy domains, we propose \textit{PolicyMoE}, a method that transforms the overall LLM into a compositional and modular system of experts with different expertise, drawing inspiration from \citep{kang2024self}. In this section, we present the details of \textit{PolicyMoE}. By dividing the policy domain expertise into three specialized expert modules—Memory Policy, Understanding Policy, and Apply Policy—\textit{PolicyMoE} creates a comprehensive system capable of handling diverse policy-related tasks with enhanced precision and efficiency.

\begin{algorithm}[ht]
\caption{Inference Procedure of \textit{PolicyMoE}}
\label{alg:policymoe}
\KwIn{Input instruction $x$}
\KwOut{Final model response $y$}
\textbf{Step 1: Expert Modules Initialization} \\
\For{each expert $i \in \{$Memory, Understanding, Application$\}$}{
    Load LoRA adapter $\Delta \Theta_i$ into base model $\Theta_0$\;
    Construct expert model $\Theta_{\text{spec}, i} = \Theta_0 + \Delta \Theta_i$\;
}
\textbf{Step 2: Routing Decision} \\
Compute routing score vector: $\mathbf{s} = \theta_r x$\;
Compute expert weights: $\boldsymbol{\alpha} = \text{softmax}(\mathbf{s})$\;
Select top-1 expert index: $i^* = \arg\max(\boldsymbol{\alpha})$\;
\textbf{Step 3: Expert Inference} \\
Use selected expert $\Theta_{\text{spec}, i^*}$ to generate response: \\
$y = \texttt{LM}_{\Theta_{\text{spec}, i^*}}(x)$\;

\Return $y$
\end{algorithm}

\subsection{Architecture Overview.}

\textit{PolicyMoE} follows the core principles of the MoE framework while specializing in policy domains:

\begin{itemize}[noitemsep,leftmargin=0pt]
    \item Expert Modules: Three dedicated expert models trained on specific policy-related capabilities:
    \begin{itemize}
        \item[$\bullet$] Memory Expert: Specializes in recalling policy facts, regulations, historical precedents, and exact policy language
        \item[$\bullet$] Understanding Expert: Focuses on interpreting policy intent, analyzing implications, and explaining policy rationales.
        \item[$\bullet$] Application Expert: Excels at applying policies to specific scenarios, predicting outcomes, and recommending implementation strategies.
    \end{itemize}
   \item Intelligent Router: A simple linear layer that is shared across all LoRA adapters, which efficiently analyzes input features to determine the most relevant policy domain expertise.
\end{itemize}

\subsection{Constructing Expert Modules} 

\textbf{Specialization with LoRA}:
Each expert module is specialized using Low-Rank Adaptation (LoRA) \citep{hu2022lora}, which introduces lightweight, trainable parameters specific to each policy domain while keeping the base LLM intact. The specialized model \( \Theta_{\text{spec}, i} \) for each domain \( i \) is defined as:
\[
\Theta_{\text{spec}, i} = \Theta_0 + \Delta \Theta_i
\]
where \( \Delta \Theta_i \) represents the LoRA parameters for domain \( i \). The forward pass for each expert module is:
\[
h_i = \theta_0 x + \theta_{B_i} \theta_{A_i} x
\]
Here, \( \theta_{B_i} \in \mathbb{R}^{d \times \text{rank}} \) and \( \theta_{A_i} \in \mathbb{R}^{\text{rank} \times k} \), with \( \text{rank} \ll \min(d, k) \).
    
\subsection{Dynamic Integration of Experts}

\textbf{Routing Mechanism}:
A router module \( \theta_r \) is introduced to analyze each input token and route it to the most appropriate expert module. The output \( h \) for each input \( x \) is computed by combining the contributions of the selected expert modules, weighted by their relevance:
\[ h = \theta_0 x + \sum_{i=1}^{n} \alpha_i \Delta \theta_i x \]
where \( \alpha \) represents the weights computed by the router:
\[ \alpha = \text{top-k}(\text{softmax}(\theta_r x)) \]
The router is trained using the aggregated synthetic data \( D = \{D_i\}_{i=1}^{n} \) to learn optimal module selection for a given task:
\[ \mathcal{L}(\theta_r) = -\mathbb{E}_{(x, y) \sim D} \left[ \log P_{\Theta_0}(y | x; \theta_r, \{\Delta \Theta_i\}_{i=1}^{n}) \right] \]

\begin{table*}[t]
\centering
\renewcommand\arraystretch{1.0}
\small
\setlength{\tabcolsep}{3pt}
\caption{Performance (accuracy) of all models on different levels and regions. Gemini-2.5, Gemini-2.0, Claude-3.5 and Claude-3.7 denote Gemini-2.5-Flash, Gemini-2.0-Flash, Claude-3.5-Sonnet and Claude-3.7-sonnet respectively. Red and blue represent the highest and lowest scores in each row respectively.}
\label{tab:main_results}
\resizebox{\textwidth}{!}{
\begin{tabular}{l l | >{\centering\arraybackslash}p{1.2cm} >{\centering\arraybackslash}p{1.2cm} >{\centering\arraybackslash}p{1.2cm} >{\centering\arraybackslash}p{1.2cm} >{\centering\arraybackslash}p{1.2cm} >{\centering\arraybackslash}p{1.2cm} >{\centering\arraybackslash}p{1.2cm} >{\centering\arraybackslash}p{1.2cm} >{\centering\arraybackslash}p{1.2cm} >{\centering\arraybackslash}p{1.2cm} >{\centering\arraybackslash}p{1.2cm}}
\toprule[1pt]
\multirow{2}{*}{\textbf{Level}} & \multirow{2}{*}{\textbf{Region}} & \textbf{GPT-4o} & \textbf{o4-mini} & \textbf{Gemini-2.5} & \textbf{Gemini-2.0} & \textbf{Claude-3.7} & \textbf{Claude-3.5} & \textbf{LLaMA-4} & \textbf{Gemma-3-27B} & \textbf{QwQ-32B} & \textbf{Deepseek-V3} & \textbf{Deepseek-R1} \\
\cmidrule(lr){3-13}
\multirow{2}{*}{Level 1} 
& CN & 46.01\% & 45.93\% & 54.06\% & 47.87\% & 55.29\% & 53.77\% & 49.81\% & \cellcolor[HTML]{ade8f4}41.75\% & 55.87\% & 48.61\% & \cellcolor[HTML]{ffccd5}62.02\% \\
& US& 52.69\% & 54.90\% & 57.73\% & 53.71\% & 58.68\% & 58.76\% & 52.55\% & 49.91\% & \cellcolor[HTML]{ade8f4}46.40\% & 50.12\% & \cellcolor[HTML]{ffccd5}59.33\% \\
\midrule
\multirow{2}{*}{Level 2} 
& CN & 56.34\% & 55.81\% & 60.57\% & 56.39\% & 60.47\% & 59.74\% & 56.56\% & 55.56\% & 59.79\% & \cellcolor[HTML]{ade8f4}55.51\% & \cellcolor[HTML]{ffccd5}62.92\% \\
& US & 63.40\% & 64.71\% & 64.91\% & 62.25\% & 68.23\% & \cellcolor[HTML]{ffccd5}68.95\% & 61.17\% & 62.17\% &\cellcolor[HTML]{ade8f4} 57.71\% & 58.62\% & 65.37\% \\
\midrule
\multirow{2}{*}{Level 3} 
& CN & 70.24\% & 79.49\% & 76.18\% & 73.80\% & 73.82\% & 72.83\% & \cellcolor[HTML]{ade8f4}68.54\% & 71.51\% & \cellcolor[HTML]{ffccd5}80.34\% & 72.33\% & 73.78\% \\
& US & 68.13\% & \cellcolor[HTML]{ffccd5}77.00\% & 69.44\% & 66.55\% & 68.28\% & 68.47\% & \cellcolor[HTML]{ade8f4}66.41\% & 68.37\% & 69.90\% & 69.39\% & 74.60\% \\
\midrule
\multirow{1}{*}{\textsc{Average}} 
&  & \ 59.47 \%& \ 62.97\% & \ 63.82\% & \ 60.10\% & \ 64.13\% & \ 63.75\% & \ 59.17\% & \cellcolor[HTML]{ade8f4} 58.21\% & \ 61.67\% & \ 59.10\% & \cellcolor[HTML]{ffccd5}66.34\% \\
\bottomrule[1pt]
\end{tabular}
}
\end{table*}

\begin{table*}[t]
\centering
\small
\renewcommand\arraystretch{1}
\setlength{\tabcolsep}{3.5pt} 
\caption{Average accuracy (\%) of all models across Chinese and US (red and blue represent the highest and lowest in each row).}
\vspace{-2pt}
\scalebox{0.92}{
\begin{tabular}{
l 
>{\centering\arraybackslash}p{1cm} >{\centering\arraybackslash}p{1.2cm} 
>{\centering\arraybackslash}p{1.2cm} >{\centering\arraybackslash}p{1.2cm}
>{\centering\arraybackslash}p{1.2cm} >{\centering\arraybackslash}p{1.2cm}
>{\centering\arraybackslash}p{1.2cm} >{\centering\arraybackslash}p{1.2cm}
>{\centering\arraybackslash}p{1.2cm} >{\centering\arraybackslash}p{1.2cm}
>{\centering\arraybackslash}p{1.2cm} >{\centering\arraybackslash}p{1.2cm}
}
\toprule[1pt]
\textbf{Region} & \textbf{GPT-4o} & \textbf{o4-mini} & \textbf{Gemini-2.5} & \textbf{Gemini-2.0} & \textbf{Claude-3.7} & \textbf{Claude-3.5} & \textbf{LLaMA-4} & \textbf{Gemma-3-27B} & \textbf{QwQ-32B} & \textbf{Deepseek-V3} & \textbf{Deepseek-R1} \\
\midrule
\textbf{Chinese} 
& 57.53\% & 60.41\% & 63.60\% & 59.35\% & 63.19\% & 62.11\% & 58.30\% & \cellcolor[HTML]{ade8f4}56.27\% & \cellcolor[HTML]{ffccd5}65.33\% & 58.82\% & 62.24\% \\
\textbf{US} 
& 61.41\% & 65.54\% & 64.03\% & 60.84\% & 65.06\% & 65.39\% & 60.04\% & 60.15\% & \cellcolor[HTML]{ade8f4}58.00\% & 59.38\% & \cellcolor[HTML]{ffccd5}66.43\% \\
\midrule
\textbf{Average} 
& 59.47\% & 62.98\% & 63.82\% & 60.10\% & 64.13\% & 63.75\% & 59.17\% & \cellcolor[HTML]{ade8f4}58.21\% & 61.67\% & 59.10\% & \cellcolor[HTML]{ffccd5}64.34\% \\
\bottomrule[1pt]
\end{tabular}
}
\vspace{-6pt}
\label{tab:average_scores}
\end{table*}

\section{Experiments}

\subsection{Setup}

\textbf{Models.} We select 11 representative state-of-art models: \texttt{GPT-4o} \citep{achiam2023gpt}, \texttt{o4-mini}, \texttt{Claude-3.7-Sonnet} \citep{anthropic_claude_3_7}, \texttt{Claude-3.5-sonnet} \citep{anthropic_claude_3_5}, \texttt{Gemini-2.5-Flash} \citep{gemini-2.5-flash}, \texttt{Gemini-2.0-Flash} \citep{gemini-2-flash}, and the open-source models: \texttt{Gemma-3-27B} \citep{team2025gemma}, \texttt{Qwen-QwQ-32B} \citep{qwen-qwq-32b}, \texttt{Llama-4} \citep{llama4}, \texttt{Deepseek-V3} \citep{liu2024deepseek} and \texttt{Deepseek-R1} \citep{guo2025deepseek}, details in \autoref{tab:models}.

\noindent\textbf{Scoring Mechanism.} \textit{PolicyBench} adopts a level-aware scoring framework tailored to question type and format.
\textbf{Levels 1–2} consist exclusively of multiple-choice and true/false questions, which are evaluated using standard accuracy: {\small \(\text{Score} = \frac{\text{\# Correct Answers}}{\text{\# Total Questions}}\)}.
\textbf{Level 3} includes both objective (multiple-choice, true/false) and subjective (open-ended) questions. Open-ended responses are scored on a 0–5 scale based on alignment with reference answers. To ensure evaluation consistency, we adopt the \textbf{LLM-as-a-Judge} \citep{zheng2023judging}: for each open-ended question, \textbf{two models are randomly sampled} from a pool of four state-of-the-art LLMs: \texttt{o4-mini}, \texttt{gemini-2.5-flash}, \texttt{claude-3.7-sonnet}, and \texttt{Deepseek-R1}, to serve as automated graders. Each grader compares the response to a reference answer and assigns a score according to predefined criteria. The final score for that question is computed as the average of the two model scores. We analyze the potential bias in \autoref{sec:bias_analysis}.

The overall Level 3 score is calculated as a weighted average across all question types:
\[
\text{Score} = \frac{S_{\text{mc}} + S_{\text{tf}} + S_{\text{oe}}}{T_{\text{mc}} + T_{\text{tf}} + 5 \times T_{\text{oe}}},
\]
where $S_{\text{mc}}, S_{\text{tf}}, S_{\text{oe}}$ denote the cumulative scores for multiple-choice, true/false, and open-ended questions, respectively, and $T_{\text{mc}}, T_{\text{tf}}, T_{\text{oe}}$ represent the corresponding question counts. The weighting reflects the maximum possible score (5) for open-ended responses. In addition, we conducted some experiments to demonstrate the robustness of LLM-as-a-Judge in \autoref{sec:judge_validation}.

\noindent\textbf{Training Setup.}
We initialize our MoE architecture using \textbf{\texttt{Qwen2.5-7B-Instruct}} as the base model, using \texttt{bfloat16} precision. Expert modules are fine-tuned using LoRA \citep{hu2022lora} with a rank of 16, scaling factor $\alpha = 32$, and dropout rate of 0.05.

Training is conducted in two stages: expert training for 3 epochs and router training for 4 epochs. We use a batch size of 4 per device with a gradient accumulation step of 4, resulting in an effective batch size of 16. The learning rate is set to $5 \times 10^{-5}$ throughout. The \textit{PolicyBench} is partitioned into an 80/20 split for training and testing. To prevent data leakage, we perform a grouped split by policy, ensuring all questions from the same source document are kept in the same set. See \autoref{sec:training} for more details.

\begin{figure*}[t]
    \centering
    \includegraphics[width=1.0\linewidth]{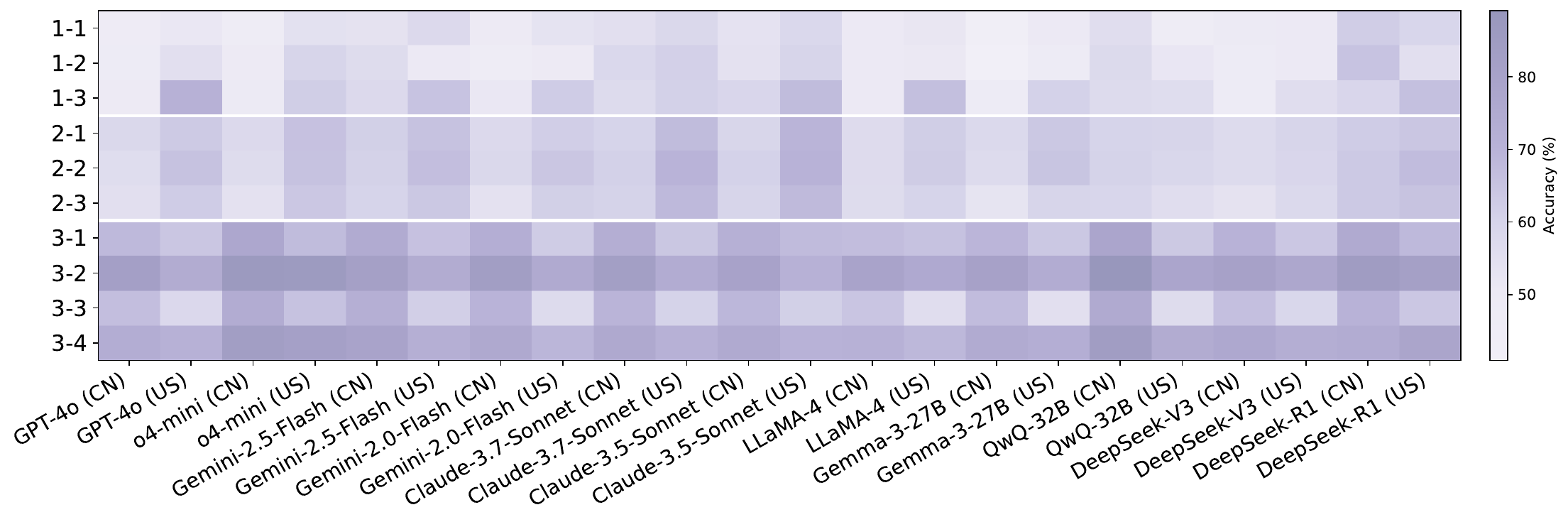}
    \caption{Model performance in 10 subtasks (ID and the specific task are shown in \autoref{tab:fined_grained_task}).}
    \label{fig:heatmap}
    \vspace{-10pt}
\end{figure*}

\subsection{Main Results}
\textbf{\textit{\ul{Performance improves from memorization to application.}}}
As shown in \autoref{tab:main_results}, models exhibit progressively higher accuracy from \textbf{Level 1} (memorization) to \textbf{Level 3} (application) across both Chinese and US settings. For instance, in the Chinese subset, average model scores range from \textbf{41.8\%–62.0\%} on Level 1 to \textbf{70.2\%–80.3\%} on Level 3. A similar trend is observed in the US subset. 
We posit that this phenomenon stems from the distinct capabilities emphasized during the different stages of LLM training. Level 1 tasks demand high-fidelity recall of specific facts (e.g., policy dates, exact terminology), which is a function of knowledge acquired during \textbf{pre-training}. While vast, this knowledge is stored implicitly, and its precise retrieval can be unreliable. In contrast, Level 2 (Understanding) and Level 3 (Application) tasks heavily rely on structured reasoning, contextual interpretation, and problem-solving—skills that are explicitly and extensively honed during \textbf{post-training} (\textit{e.g., instruction tuning, RLHF \citep{ouyang2022training}}). Therefore, the models' superior performance on these more complex tasks likely reflects a stronger alignment with the generalizable reasoning abilities optimized during fine-tuning, rather than a deeper mastery of the policy domain itself.
\textit{Among all evaluated models, \texttt{DeepSeek-R1}} achieves the highest overall accuracy at \textbf{66.34\%}, making it a strong candidate for practitioners seeking a general-purpose model for policy-related applications.

\noindent\textbf{\textit{\ul{Models excel at structured reasoning tasks but falter on abstract concepts.}}}
As shown in the task-wise heatmap (\autoref{fig:heatmap}), models consistently achieve higher accuracy on specific application-oriented tasks, particularly \textbf{Policy-Based Numerical Reasoning} and \textbf{Scenario-Based Decision-Making}. For many models, accuracy in these categories exceeds \textbf{75\%}, with some surpassing \textbf{80\%}. These tasks typically involve concrete conditions, rule-based logic, or everyday reasoning scenarios—formats that closely align with the pretraining and instruction-following capabilities of large language models. In contrast, accuracy remains relatively lower on more abstract or ambiguous tasks such as \textbf{Ideas} and \textbf{Institutions}, which require understanding latent policy concepts or institutional relationships. This suggests that LLMs are better equipped to handle tasks with clear logical structures than those requiring interpretive or conceptual comprehension.

\noindent\textbf{\textit{\ul{Models consistently perform better on US policy questions than Chinese ones.}}}
As shown in \autoref{tab:average_scores}, most models achieve higher accuracy on US policy questions than on their Chinese counterparts. The overall average improves from \textbf{61.02\%} (CN) to \textbf{62.39\%} (US), with models like \texttt{o4-mini} rising from \textbf{60.41\%} to \textbf{65.54\%}, and \texttt{Claude-3.5-Sonnet} from \textbf{62.11\%} to \textbf{65.39\%}. An exception is \texttt{QwQ-32B}, which performs notably better in Chinese (\textbf{65.33\%}) than in English (\textbf{58.00\%}). This overall trend may be attributed to the dominance of English in pretraining corpora, as well as the higher syntactic and semantic density of Chinese policy texts. These results highlight the need for more robust cross-lingual policy understanding in LLMs. Further more, we also conduct an error analysis (detailed in \autoref{sec:error_study}).

\begin{table}[htbp]
\centering
\caption{Qwen2.5-7B-Instruct performance across levels and regions before and after training (\%)}
\vspace{-5pt}
\label{tab:training_horizontal}
\renewcommand{\arraystretch}{1}
\resizebox{0.95\linewidth}{!}{
\begin{tabular}{cc | c c c}
\toprule[1pt]
\textbf{Level} & \textbf{Region} & \textbf{Original} & \textbf{Training} & \textbf{$\Delta$} \\
\midrule
\multirow{2}{*}{\textbf{Level 1}} & CN & 36.85\% & 41.83\% & \textcolor{green!50!black}{$\uparrow$~13.51\%} \\
                         & US & 23.35\% & 35.43\% & \textcolor{green!50!black}{$\uparrow$~51.73\%} \\
\midrule
\multirow{2}{*}{\textbf{Level 2}} & CN & 45.68\% & 47.02\% & \textcolor{green!50!black}{$\uparrow$~2.93\%} \\
                         & US & 42.31\% & 42.78\% & \textcolor{green!50!black}{$\uparrow$~1.11\%} \\
\midrule
\multirow{2}{*}{\textbf{Level 3}} & CN & 64.73\% & 69.12\% & \textcolor{green!50!black}{$\uparrow$~6.78\%} \\
                         & US & 46.65\% & 57.48\% & \textcolor{green!50!black}{$\uparrow$~23.22\%} \\
\bottomrule[1pt]
\end{tabular}
}
\vspace{-5pt}
\end{table}

\vspace{-10pt}
\subsection{Results of \textit{PolicyMoE}}

\autoref{tab:training_horizontal} reports the performance of our model before and after fine-tuning with the \textit{PolicyMoE} framework. The goal of this experiment is not to pursue state-of-the-art results with a moderately sized base model (Qwen2.5-7B-Instruct), but to demonstrate the efficacy of our expert specialization approach. Accordingly, we emphasize the \textbf{relative improvements} achieved. Notably, the fine-tuned 7B model not only shows substantial gains but also outperforms several larger baselines in \autoref{tab:main_results}, underscoring the effectiveness of domain-specific adaptation.

Across task levels, performance improves consistently after fine-tuning. The largest gain appears in US Level 1 tasks, where accuracy rises from 23.35\% to 35.43\%—a relative improvement of over 50\%. China also records a 13.51\% gain at the same level, highlighting the benefit of injecting structured domain knowledge. Improvements on Level 2, which emphasizes policy comprehension, are more modest (2.93\% for China and 1.11\% for the US), suggesting that higher-level reasoning is less sensitive to task-specific fine-tuning and may require advanced strategies such as chain-of-thought prompting \cite{wei2022chain} or richer supervision. By contrast, Level 3 tasks show more pronounced gains, with the US domain achieving a 23.22\% relative improvement, indicating that the Application Expert effectively supports contextual reasoning and scenario-based decision-making.

Overall, \textit{PolicyMoE} delivers clear benefits for both factual recall and applied reasoning. While improvements in abstract comprehension remain limited, the results point to a promising path toward specialized, capable models without relying on prohibitively large architectures. We also compare \textit{PolicyMoE} with standard LoRA \cite{hu2022lora} in \autoref{sec:ablation}.

\begin{figure}[t]
    \centering
    \includegraphics[width=1.0\linewidth]{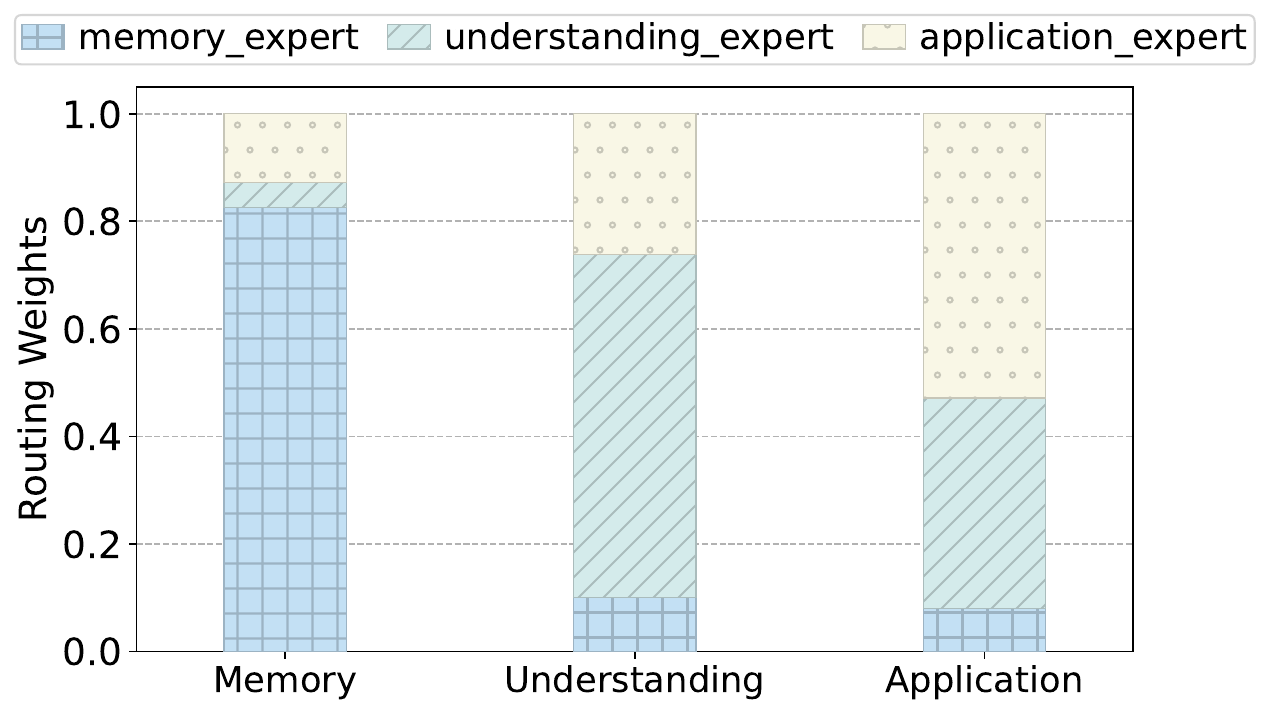}
    \vspace{-10pt}
    \caption{Routing distributions over three experts for each level.}
    \vspace{-10pt}
    \label{fig:router}
\end{figure}

\begin{figure}
    \centering
    \includegraphics[width=0.95\linewidth]{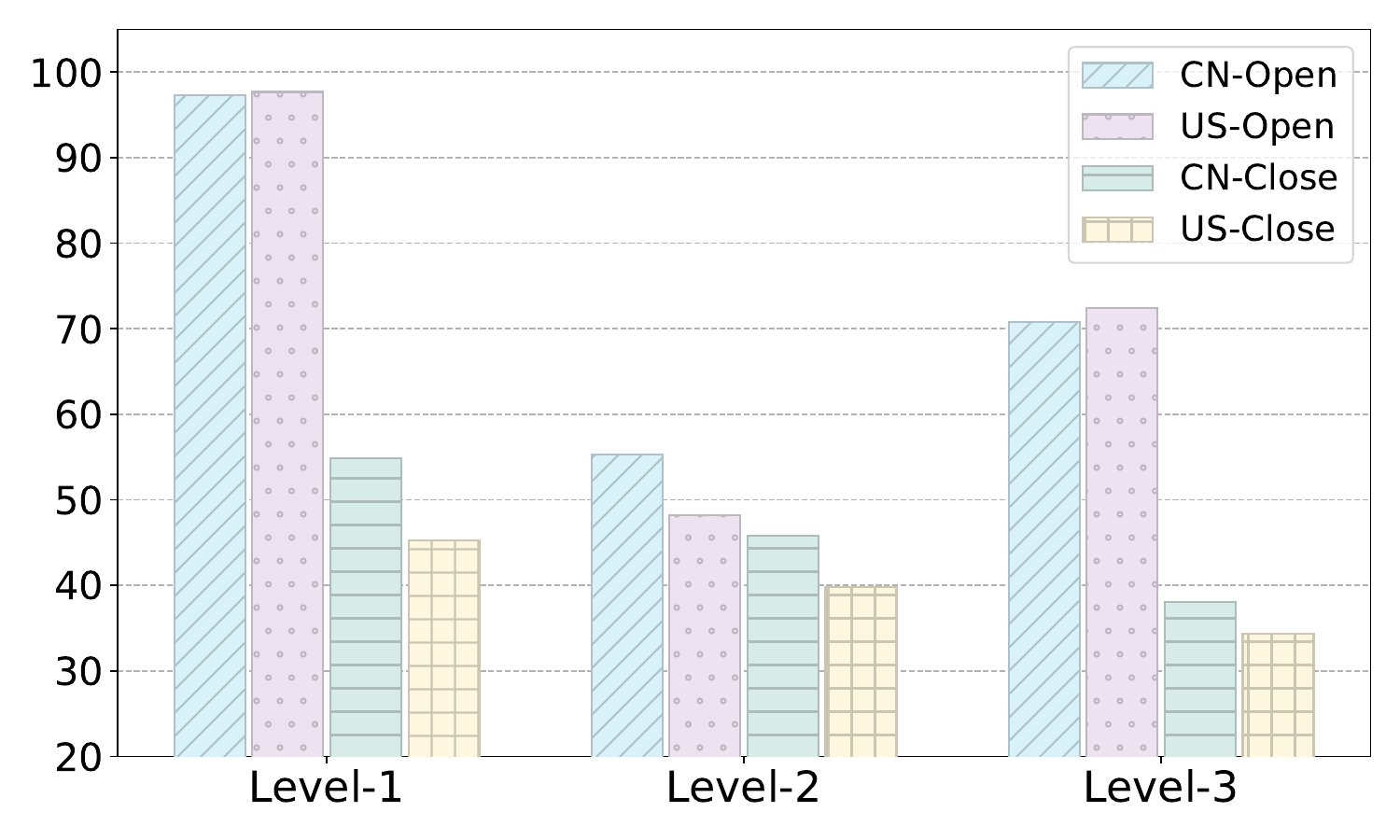}
    \caption{Human accuracy on three levels (\%).}
    \vspace{-15pt}
    \label{fig:human_performance}
\end{figure}

\subsection{\textit{PolicyMoE} Router Analysis}

To analyze the behavior of the router module, we randomly sample 10 questions from each cognitive level (Level 1–3) and compute the average expert weights assigned by the router. This allows us to observe how the model distributes attention across three experts in response to different types of tasks.

As shown in \autoref{fig:router}, router weight patterns show clear specialization for factual tasks, with memory weights peaking over 80\% when the memory expert is selected. In contrast, understanding and application selections result in more distributed weights, reflecting the multi-dimensional nature of these tasks and the need for shared reasoning across modules. 

\subsection{Human Performance}

To contextualize LLM performance, we established human baselines with 12 university students from the United States and China, distinct from the annotators. Participants, proficient in English or Mandarin but without policy expertise, each answered 100 randomly sampled \textit{PolicyBench} questions under two conditions: \textbf{open-book} (with access to policy texts) and \textbf{closed-book} (relying on prior knowledge). As shown in \autoref{fig:human_performance}, the results indicate that:

\begin{itemize}[noitemsep,leftmargin=0pt]
\item \textbf{Open- vs. closed-book gaps distinguish memory from reasoning.} The large gap at Level 1 reflects reliance on factual recall, while the minimal difference at Level 2 suggests a shift toward reasoning-intensive challenges.
\item \textbf{Cross-linguistic consistency supports benchmark validity.} Comparable performance across languages within each setting indicates that task difficulty is not driven by language differences.
\item \textbf{Non-monotonic accuracy across levels.} Accuracy drops at Level 2 and partially recovers at Level 3, suggesting that abstract reasoning without direct retrieval is most challenging, whereas Level 3 allows participants to combine reasoning with general knowledge.
\end{itemize}



\vspace{-0.1in}
\section{Conclusion}
\vspace{-0.05in}

We introduce \textit{PolicyBench}, a cross-system benchmark (US-China) assessing LLM comprehension of public policy, which reveals that models perform well on recall and application tasks, they struggle with understanding questions involving policy intent and institutional reasoning. To bridge this gap, we propose \textit{PolicyMoE}, a domain-specialized MoE model that achieves improved performance. Our findings underscore the need for targeted adaptation to support real-world policy analysis.

\section*{Limitations}
\label{sec:lim}

\textbf{Geographic and systemic scope.} \textit{PolicyBench} currently covers only China and the United States. While these two cases offer a strong contrast across governance systems, they cannot fully represent the diversity of global policy environments. Extending to additional regions would improve generalizability and cross-cultural robustness.

\noindent\textbf{Task diversity.} The benchmark mainly relies on multiple-choice and true/false formats, with limited coverage of open-ended tasks. Real-world policy analysis, however, involves greater ambiguity, nuance, and value-sensitive reasoning than what structured formats can capture. Expanding task types would better reflect such challenges.

\noindent\textbf{Model adaptation and architecture.} \textit{PolicyMoE} shows clear improvements in factual recall and applied reasoning, but its gains in abstract comprehension remain limited. Moreover, the current router design can only select one expert per query, whereas increasingly complex tasks may require activating multiple experts simultaneously. Future work could explore reasoning-focused models, advanced prompting strategies, and more flexible routing mechanisms that allocate experts based on learned weight distributions.

\section*{Ethics Statement}

All data used in this study were collected from publicly accessible platforms in compliance with ethical and legal standards. No proprietary or private materials were included. For the human evaluation component, all participants were recruited on a voluntary basis and provided informed consent prior to their involvement. They were clearly informed of the research purpose and their rights, including the right to withdraw at any stage. Finally, all content generated by large language models was carefully reviewed to ensure that it did not contain sensitive, harmful, or inappropriate material. The study adheres to responsible AI research practices and aims to contribute to the safe and transparent development of policy-focused benchmarks and models.

\bibliography{custom}

@article{vaswani2017attention,
  title={Attention is all you need},
  author={Vaswani, Ashish and Shazeer, Noam and Parmar, Niki and Uszkoreit, Jakob and Jones, Llion and Gomez, Aidan N and Kaiser, {\L}ukasz and Polosukhin, Illia},
  journal={Advances in neural information processing systems},
  volume={30},
  year={2017}
}

@article{achiam2023gpt,
  title={Gpt-4 technical report},
  author={Achiam, Josh and Adler, Steven and Agarwal, Sandhini and Ahmad, Lama and Akkaya, Ilge and Aleman, Florencia Leoni and Almeida, Diogo and Altenschmidt, Janko and Altman, Sam and Anadkat, Shyamal and others},
  journal={arXiv preprint arXiv:2303.08774},
  year={2023}
}

@article{touvron2023llama,
  title={Llama: Open and efficient foundation language models},
  author={Touvron, Hugo and Lavril, Thibaut and Izacard, Gautier and Martinet, Xavier and Lachaux, Marie-Anne and Lacroix, Timoth{\'e}e and Rozi{\`e}re, Baptiste and Goyal, Naman and Hambro, Eric and Azhar, Faisal and others},
  journal={arXiv preprint arXiv:2302.13971},
  year={2023}
}

@inproceedings{yuan2022wordcraft,
  title={Wordcraft: story writing with large language models},
  author={Yuan, Ann and Coenen, Andy and Reif, Emily and Ippolito, Daphne},
  booktitle={Proceedings of the 27th International Conference on Intelligent User Interfaces},
  pages={841--852},
  year={2022}
}

@inproceedings{zhu2024multilingual,
  title={Multilingual Machine Translation with Large Language Models: Empirical Results and Analysis},
  author={Zhu, Wenhao and Liu, Hongyi and Dong, Qingxiu and Xu, Jingjing and Huang, Shujian and Kong, Lingpeng and Chen, Jiajun and Li, Lei},
  booktitle={Findings of the Association for Computational Linguistics: NAACL 2024},
  pages={2765--2781},
  year={2024}
}

@inproceedings{svyatkovskiy2020intellicode,
  title={Intellicode compose: Code generation using transformer},
  author={Svyatkovskiy, Alexey and Deng, Shao Kun and Fu, Shengyu and Sundaresan, Neel},
  booktitle={Proceedings of the 28th ACM joint meeting on European software engineering conference and symposium on the foundations of software engineering},
  pages={1433--1443},
  year={2020}
}

@inproceedings{wangglue,
  title={GLUE: A Multi-Task Benchmark and Analysis Platform for Natural Language Understanding},
  author={Wang, Alex and Singh, Amanpreet and Michael, Julian and Hill, Felix and Levy, Omer and Bowman, Samuel R},
  booktitle={International Conference on Learning Representations}
}

@article{xiang2023beyond,
  title={Beyond the Imitation Game: Collaborative benchmark for measuring and extrapolating the capabilities of language models [co-authored]: What is the Tao?},
  author={Xiang, Alice},
  journal={Transactions on Machine Learning Research},
  year={2023}
}

@inproceedings{joshi2017triviaqa,
  title={TriviaQA: A Large Scale Distantly Supervised Challenge Dataset for Reading Comprehension},
  author={Joshi, Mandar and Choi, Eunsol and Weld, Daniel S and Zettlemoyer, Luke},
  booktitle={Proceedings of the 55th Annual Meeting of the Association for Computational Linguistics (Volume 1: Long Papers)},
  pages={1601--1611},
  year={2017}
}

@inproceedings{fei2024lawbench,
  title={LawBench: Benchmarking Legal Knowledge of Large Language Models},
  author={Fei, Zhiwei and Shen, Xiaoyu and Zhu, Dawei and Zhou, Fengzhe and Han, Zhuo and Huang, Alan and Zhang, Songyang and Chen, Kai and Yin, Zhixin and Shen, Zongwen and others},
  booktitle={Proceedings of the 2024 Conference on Empirical Methods in Natural Language Processing},
  pages={7933--7962},
  year={2024}
}

@article{guha2023legalbench,
  title={Legalbench: A collaboratively built benchmark for measuring legal reasoning in large language models},
  author={Guha, Neel and Nyarko, Julian and Ho, Daniel and R{\'e}, Christopher and Chilton, Adam and Chohlas-Wood, Alex and Peters, Austin and Waldon, Brandon and Rockmore, Daniel and Zambrano, Diego and others},
  journal={Advances in Neural Information Processing Systems},
  volume={36},
  pages={44123--44279},
  year={2023}
}

@article{kang2024self,
  title={Self-moe: Towards compositional large language models with self-specialized experts},
  author={Kang, Junmo and Karlinsky, Leonid and Luo, Hongyin and Wang, Zhen and Hansen, Jacob and Glass, James and Cox, David and Panda, Rameswar and Feris, Rogerio and Ritter, Alan},
  journal={arXiv preprint arXiv:2406.12034},
  year={2024}
}

@article{zhou2024lawgpt,
  title={LawGPT: A Chinese Legal Knowledge-Enhanced Large Language Model},
  author={Zhou, Zhi and Shi, Jiang-Xin and Song, Peng-Xiao and Yang, Xiaowen and Jin, Yi-Xuan and Guo, Lan-Zhe and Li, Yu-Feng},
  journal={CoRR},
  year={2024}
}

@article{krathwohl2002revision,
  title={A revision of Bloom's taxonomy: An overview},
  author={Krathwohl, David R},
  journal={Theory into practice},
  volume={41},
  number={4},
  pages={212--218},
  year={2002},
  publisher={Taylor \& Francis}
}

@article{hall1996political,
  title={Political science and the three new institutionalisms},
  author={Hall, Peter A and Taylor, Rosemary CR},
  journal={Political studies},
  volume={44},
  number={5},
  pages={936--957},
  year={1996},
  publisher={SAGE Publications Sage UK: London, England}
}

@article{bao2024autobench,
  title={AutoBench-V: Can Large Vision-Language Models Benchmark Themselves?},
  author={Bao, Han and Huang, Yue and Wang, Yanbo and Ye, Jiayi and Wang, Xiangqi and Chen, Xiuying and Zhao, Yue and Zhou, Tianyi and Elhoseiny, Mohamed and Zhang, Xiangliang},
  journal={arXiv preprint arXiv:2410.21259},
  year={2024}
}

@inproceedings{wang2024mmlu,
  title={Mmlu-pro: A more robust and challenging multi-task language understanding benchmark},
  author={Wang, Yubo and Ma, Xueguang and Zhang, Ge and Ni, Yuansheng and Chandra, Abhranil and Guo, Shiguang and Ren, Weiming and Arulraj, Aaran and He, Xuan and Jiang, Ziyan and others},
  booktitle={The Thirty-eight Conference on Neural Information Processing Systems Datasets and Benchmarks Track},
  year={2024}
}

@inproceedings{xiao2023evaluating,
  title={Evaluating reading comprehension exercises generated by LLMs: A showcase of ChatGPT in education applications},
  author={Xiao, Changrong and Xu, Sean Xin and Zhang, Kunpeng and Wang, Yufang and Xia, Lei},
  booktitle={Proceedings of the 18th workshop on innovative use of NLP for building educational applications (BEA 2023)},
  pages={610--625},
  year={2023}
}

@article{tang2023does,
  title={Does synthetic data generation of llms help clinical text mining?},
  author={Tang, Ruixiang and Han, Xiaotian and Jiang, Xiaoqian and Hu, Xia},
  journal={arXiv preprint arXiv:2303.04360},
  year={2023}
}

@article{pesch2025potentials,
  title={Potentials and Challenges of Large Language Models (LLMs) in the Context of Administrative Decision-Making},
  author={Pesch, Paulina Jo},
  journal={European Journal of Risk Regulation},
  pages={1--20},
  year={2025},
  publisher={Cambridge University Press}
}

@article{hou2025urban,
  title={Urban sensing in the era of large language models},
  author={Hou, Ce and Zhang, Fan and Li, Yong and Li, Haifeng and Mai, Gengchen and Kang, Yuhao and Yao, Ling and Yu, Wenhao and Yao, Yao and Gao, Song and others},
  journal={The Innovation},
  volume={6},
  number={1},
  year={2025},
  publisher={Elsevier}
}

@article{hu2022lora,
  title={Lora: Low-rank adaptation of large language models.},
  author={Hu, Edward J and Shen, Yelong and Wallis, Phillip and Allen-Zhu, Zeyuan and Li, Yuanzhi and Wang, Shean and Wang, Lu and Chen, Weizhu and others},
  journal={ICLR},
  volume={1},
  number={2},
  pages={3},
  year={2022}
}

@article{jacobs1991adaptive,
  title={Adaptive mixtures of local experts},
  author={Jacobs, Robert A and Jordan, Michael I and Nowlan, Steven J and Hinton, Geoffrey E},
  journal={Neural computation},
  volume={3},
  number={1},
  pages={79--87},
  year={1991},
  publisher={MIT Press}
}

@article{jordan1994hierarchical,
  title={Hierarchical mixtures of experts and the EM algorithm},
  author={Jordan, Michael I and Jacobs, Robert A},
  journal={Neural computation},
  volume={6},
  number={2},
  pages={181--214},
  year={1994},
  publisher={MIT Press}
}

@article{chen2021evaluating,
  title={Evaluating large language models trained on code},
  author={Chen, Mark and Tworek, Jerry and Jun, Heewoo and Yuan, Qiming and Pinto, Henrique Ponde De Oliveira and Kaplan, Jared and Edwards, Harri and Burda, Yuri and Joseph, Nicholas and Brockman, Greg and others},
  journal={arXiv preprint arXiv:2107.03374},
  year={2021}
}

@article{cobbe2021training,
  title={Training verifiers to solve math word problems},
  author={Cobbe, Karl and Kosaraju, Vineet and Bavarian, Mohammad and Chen, Mark and Jun, Heewoo and Kaiser, Lukasz and Plappert, Matthias and Tworek, Jerry and Hilton, Jacob and Nakano, Reiichiro and others},
  journal={arXiv preprint arXiv:2110.14168},
  year={2021}
}

@misc{baek2025researchagentiterativeresearchidea,
      title={ResearchAgent: Iterative Research Idea Generation over Scientific Literature with Large Language Models}, 
      author={Jinheon Baek and Sujay Kumar Jauhar and Silviu Cucerzan and Sung Ju Hwang},
      year={2025},
      eprint={2404.07738},
      archivePrefix={arXiv},
      primaryClass={cs.CL},
      url={https://arxiv.org/abs/2404.07738}, 
}

@misc{asai2024openscholarsynthesizingscientificliterature,
      title={OpenScholar: Synthesizing Scientific Literature with Retrieval-augmented LMs}, 
      author={Akari Asai and Jacqueline He and Rulin Shao and Weijia Shi and Amanpreet Singh and Joseph Chee Chang and Kyle Lo and Luca Soldaini and Sergey Feldman and Mike D'arcy and David Wadden and Matt Latzke and Minyang Tian and Pan Ji and Shengyan Liu and Hao Tong and Bohao Wu and Yanyu Xiong and Luke Zettlemoyer and Graham Neubig and Dan Weld and Doug Downey and Wen-tau Yih and Pang Wei Koh and Hannaneh Hajishirzi},
      year={2024},
      eprint={2411.14199},
      archivePrefix={arXiv},
      primaryClass={cs.CL},
      url={https://arxiv.org/abs/2411.14199}, 
}

@misc{he2025pasallmagentcomprehensive,
      title={PaSa: An LLM Agent for Comprehensive Academic Paper Search}, 
      author={Yichen He and Guanhua Huang and Peiyuan Feng and Yuan Lin and Yuchen Zhang and Hang Li and Weinan E},
      year={2025},
      eprint={2501.10120},
      archivePrefix={arXiv},
      primaryClass={cs.IR},
      url={https://arxiv.org/abs/2501.10120}, 
}

@misc{colombo2024saullm7bpioneeringlargelanguage,
      title={SaulLM-7B: A pioneering Large Language Model for Law}, 
      author={Pierre Colombo and Telmo Pessoa Pires and Malik Boudiaf and Dominic Culver and Rui Melo and Caio Corro and Andre F. T. Martins and Fabrizio Esposito and Vera Lúcia Raposo and Sofia Morgado and Michael Desa},
      year={2024},
      eprint={2403.03883},
      archivePrefix={arXiv},
      primaryClass={cs.CL},
      url={https://arxiv.org/abs/2403.03883}, 
}

@inproceedings{Savelka_2023, series={ICAIL 2023},
   title={Unlocking Practical Applications in Legal Domain: Evaluation of GPT for Zero-Shot Semantic Annotation of Legal Texts},
   url={http://dx.doi.org/10.1145/3594536.3595161},
   DOI={10.1145/3594536.3595161},
   booktitle={Proceedings of the Nineteenth International Conference on Artificial Intelligence and Law},
   publisher={ACM},
   author={Savelka, Jaromir},
   year={2023},
   month=jun, pages={447–451},
   collection={ICAIL 2023} }

@article{wang2023docllm,
  title={DocLLM: A layout-aware generative language model for multimodal document understanding},
  author={Wang, Dongsheng and Raman, Natraj and Sibue, Mathieu and Ma, Zhiqiang and Babkin, Petr and Kaur, Simerjot and Pei, Yulong and Nourbakhsh, Armineh and Liu, Xiaomo},
  journal={arXiv preprint arXiv:2401.00908},
  year={2023}
}

@article{tsatsaronis2015overview,
  title={An overview of the BIOASQ large-scale biomedical semantic indexing and question answering competition},
  author={Tsatsaronis, George and Balikas, Georgios and Malakasiotis, Prodromos and Partalas, Ioannis and Zschunke, Matthias and Alvers, Michael R and Weissenborn, Dirk and Krithara, Anastasia and Petridis, Sergios and Polychronopoulos, Dimitris and others},
  journal={BMC bioinformatics},
  volume={16},
  pages={1--28},
  year={2015},
  publisher={Springer}
}

@article{jin2021disease,
  title={What disease does this patient have? a large-scale open domain question answering dataset from medical exams},
  author={Jin, Di and Pan, Eileen and Oufattole, Nassim and Weng, Wei-Hung and Fang, Hanyi and Szolovits, Peter},
  journal={Applied Sciences},
  volume={11},
  number={14},
  pages={6421},
  year={2021},
  publisher={MDPI}
}

@inproceedings{yang2015wikiqa,
  title={Wikiqa: A challenge dataset for open-domain question answering},
  author={Yang, Yi and Yih, Wen-tau and Meek, Christopher},
  booktitle={Proceedings of the 2015 conference on empirical methods in natural language processing},
  pages={2013--2018},
  year={2015}
}

@article{team2025gemma,
  title={Gemma 3 technical report},
  author={Team, Gemma and Kamath, Aishwarya and Ferret, Johan and Pathak, Shreya and Vieillard, Nino and Merhej, Ramona and Perrin, Sarah and Matejovicova, Tatiana and Ram{\'e}, Alexandre and Rivi{\`e}re, Morgane and others},
  journal={arXiv preprint arXiv:2503.19786},
  year={2025}
}

@article{liu2024deepseek,
  title={Deepseek-v3 technical report},
  author={Liu, Aixin and Feng, Bei and Xue, Bing and Wang, Bingxuan and Wu, Bochao and Lu, Chengda and Zhao, Chenggang and Deng, Chengqi and Zhang, Chenyu and Ruan, Chong and others},
  journal={arXiv preprint arXiv:2412.19437},
  year={2024}
}

@article{guo2025deepseek,
  title={Deepseek-r1: Incentivizing reasoning capability in llms via reinforcement learning},
  author={Guo, Daya and Yang, Dejian and Zhang, Haowei and Song, Junxiao and Zhang, Ruoyu and Xu, Runxin and Zhu, Qihao and Ma, Shirong and Wang, Peiyi and Bi, Xiao and others},
  journal={arXiv preprint arXiv:2501.12948},
  year={2025}
}

@online{anthropic_claude_3_7,
    title={Anthropic claude-3.7-sonnet},
    author={Anthropic},
    year={2025},
    note={https://www.anthropic.com/claude/sonnet}
}

@online{llama4,
    title={Meta Llama-4},
    author={Meta},
    year={2025},
    note={https://www.llama.com/models/llama-4/}
}

@online{gemini-2.5-flash,
    title={Google Gemini-2.5-Flash},
    author={Google},
    year={2025},
    note={https://deepmind.google/technologies/gemini/flash/}
}

@online{gemini-2-flash,
    title={Google Gemini-2-Flash},
    author={Google},
    year={2024},
    note={https://deepmind.google/technologies/gemini/flash-lite/}
}

@online{anthropic_claude_3_5,
    title={Anthropic claude-3.5-sonnet},
    author={Anthropic},
    year={2024},
    note={https://www.anthropic.com/news/claude-3-5-sonnet}
}

@online{qwen-qwq-32b,
    title={Qwen-QwQ-32B},
    author={QwQ},
    year={2024},
    note={https://qwenlm.github.io/blog/qwq-32b/}
}

@article{zheng2023judging,
  title={Judging llm-as-a-judge with mt-bench and chatbot arena},
  author={Zheng, Lianmin and Chiang, Wei-Lin and Sheng, Ying and Zhuang, Siyuan and Wu, Zhanghao and Zhuang, Yonghao and Lin, Zi and Li, Zhuohan and Li, Dacheng and Xing, Eric and others},
  journal={Advances in Neural Information Processing Systems},
  volume={36},
  pages={46595--46623},
  year={2023}
}

@article{wei2022chain,
  title={Chain-of-thought prompting elicits reasoning in large language models},
  author={Wei, Jason and Wang, Xuezhi and Schuurmans, Dale and Bosma, Maarten and Xia, Fei and Chi, Ed and Le, Quoc V and Zhou, Denny and others},
  journal={Advances in neural information processing systems},
  volume={35},
  pages={24824--24837},
  year={2022}
}

@article{kang2025policysimeval,
  title={PolicySimEval: A Benchmark for Evaluating Policy Outcomes through Agent-Based Simulation},
  author={Kang, Jiaju and Han, Puyu and Zhang, Tian and Gong, Luqi},
  journal={arXiv preprint arXiv:2502.07853},
  year={2025}
}

@article{liang2025benchmarking,
  title={Benchmarking llms for political science: A united nations perspective},
  author={Liang, Yueqing and Yang, Liangwei and Wang, Chen and Xia, Congying and Meng, Rui and Xu, Xiongxiao and Wang, Haoran and Payani, Ali and Shu, Kai},
  journal={arXiv preprint arXiv:2502.14122},
  year={2025}
}

@article{safaei2024end,
  title={The end of the policy analyst? Testing the capability of artificial intelligence to generate plausible, persuasive, and useful policy analysis},
  author={Safaei, Mehrdad and Longo, Justin},
  journal={Digital Government: Research and Practice},
  volume={5},
  number={1},
  pages={1--35},
  year={2024},
  publisher={ACM New York, NY}
}

@article{karacapilidis2024generative,
  title={Generative AI and Public Deliberation: A Framework for LLM-augmented Digital Democracy},
  author={Karacapilidis, Nikos and Kalampokis, Evangelos and Giarelis, Nikolaos and Mastrokostas, Charalampos},
  journal={Proceedings http://ceur-ws. org ISSN},
  volume={1613},
  pages={0073},
  year={2024}
}

@article{ouyang2022training,
  title={Training language models to follow instructions with human feedback},
  author={Ouyang, Long and Wu, Jeffrey and Jiang, Xu and Almeida, Diogo and Wainwright, Carroll and Mishkin, Pamela and Zhang, Chong and Agarwal, Sandhini and Slama, Katarina and Ray, Alex and others},
  journal={Advances in neural information processing systems},
  volume={35},
  pages={27730--27744},
  year={2022}
}

@article{zar2005spearman,
  title={Spearman rank correlation},
  author={Zar, Jerrold H},
  journal={Encyclopedia of biostatistics},
  volume={7},
  year={2005},
  publisher={Wiley Online Library}
}

@article{bao2026position,
  title={Position: General Alignment Has Hit a Ceiling; Edge Alignment Must Be Taken Seriously},
  author={Bao, Han and Huang, Yue and Wang, Xiaoda and Zhang, Zheyuan and Zhou, Yujun and Yang, Carl and Zhang, Xiangliang and Ye, Yanfang},
  journal={arXiv preprint arXiv:2602.20042},
  year={2026}
}

\appendix

\clearpage
\section{Details of Data Collection}
\label{sec:data}

This section details the curation of our dataset, outlining the collection methodology for both policy documents and their supplementary materials, the tools utilized in the process, and the filtering pipeline applied to ensure data quality.

\subsection{Data Collection Tools and Process}

Our data was collected primarily between January 2015 and March 2025 using a hybrid approach to navigate the structural inconsistencies of official government websites.

The majority of documents were collected manually from the portals listed in \autoref{tab:data_sources}. This manual approach was essential for bypassing complex site navigation and security mechanisms, and for ensuring the correct retrieval of policy documents with associated attachments, which challenge standard web scrapers. For a small subset of highly structured content, such as news archives, we employed targeted automation scripts using Python's \textbf{Selenium} library to assist with batch-downloading.

\subsection{Data Filtering and Curation Pipeline}

To ensure the quality and relevance of our final dataset, all collected documents underwent a rigorous multi-stage filtering and curation pipeline. The objective was to create a clean, non-redundant, and substantive corpus for constructing \textbf{PolicyBench}. The pipeline consisted of the following sequential steps:

\begin{enumerate}
    \item \textbf{Duplicate Removal:} The first step was to eliminate redundant files. We identified and removed duplicates by checking for high similarity in document titles and textual content. Initially, documents with identical or near-identical titles were flagged, after which their content overlap was assessed. Documents with a high degree of textual similarity were considered duplicates, and all but the most complete version were discarded.

    \item \textbf{Substantive Content Filtering:} Next, we filtered out documents that were not substantive policy texts. A document was classified as "non-substantive" and excluded if it met any of the following criteria:
    \begin{itemize}
        \item It was purely administrative or procedural (e.g., public meeting announcements, personnel appointment notices, holiday schedules).
        \item It was a table of contents, an index, or a cover page without the corresponding full document.
        \item The document's title contained keywords from a predefined exclusion list, such as \texttt{"Notice of Public Hearing"}, \texttt{"Personnel Appointments"}, \texttt{"Weekly Agenda"}, or \texttt{"Annual Report Summary"}.
    \end{itemize}

    \item \textbf{Temporal and Relevancy Filtering:} Finally, we applied a filter to remove documents that were considered "outdated" or irrelevant to the contemporary policy landscape. A policy was flagged and removed if:
    \begin{itemize}
        \item It was explicitly superseded by a more recent version or subsequent legislation from the same issuing authority.
        \item It was promulgated before the year 2000, which we established as the historical cutoff for our benchmark to maintain modern relevance.
    \end{itemize}
\end{enumerate}

This structured pipeline allowed us to distill the large volume of collected data into the high-quality, curated set of 721 Chinese policies and 603 US policies that form the foundation of \textit{\textbf{PolicyBench}}.

\section{Details of Experiment Setting}
\label{sec:setting}

\subsection{Details of \textit{PolicyBench}}
\label{sec:details_policybench}

\textbf{Level-1: Memorization Task.} Level-1 tasks are designed to assess factual recall. We begin by using large language models to automatically generate cloze-style and true/false questions. Cloze questions are created by masking factual elements in policy texts—such as \textit{dates}, \textit{legal terms}, \textit{organization names}, and \textit{key definitions}—that reflect domain-specific knowledge and span various policy areas. These cloze items are then transformed into multiple-choice questions. To construct high-quality distractors, we prompt multiple LLMs independently to generate alternative options inspired by \citep{bao2024autobench}. This multi-model strategy reduces single-model bias and increases the plausibility and diversity of distractors, thereby enhancing the benchmark's robustness.

\textbf{Level-2: Understanding Task.} Level-2 tasks evaluate a model's ability to comprehend the deeper meaning and context of policy content. We prompt LLMs to analyze policies using the 3I framework from policy studies, which highlights three core dimensions: \textit{Ideas} (underlying beliefs and values), \textit{Interests} (stakeholders involved), and \textit{Institutions} (rules and structures guiding implementation) \citep{hall1996political}. Based on these structured analyses, we generate question-answer pairs that probe a model's understanding of policy motivations, actors, and institutional dynamics. The question generation process parallels that of Level-1: we first construct cloze-style prompts grounded in 3I insights, then convert them into multiple-choice format with distractors generated via multiple LLMs to improve quality and difficulty.

\textbf{Level-3: Application Task.} Level-3 tasks focus on practical reasoning and real-world contextual adaptation. To build these tasks, we draw on supplementary policy materials (\textit{e.g., official commentaries, media coverage, expert interviews, and public consultations}) to develop realistic scenarios where a policy might be applied. Based on these scenarios, we recruit students with relevant academic backgrounds to manually craft questions that require reasoning about a policy’s implications, suitability, or potential outcomes in novel contexts. This manual, context-driven approach ensures that Level-3 tasks closely mirror real-world decision-making challenges and reflect authentic policy discourse.

\textbf{Expert-Led Question Design for Levels 2 and 3.} To ensure high cognitive alignment and domain validity, the construction of \textbf{Level 2} (Understanding) and \textbf{Level 3} (Application) items was strictly led and executed by senior domain experts. The core writing team consisted of five senior Ph.D. candidates specializing in Public Policy, Law, and Computational Social Science. While three undergraduate assistants provided support for data formatting and preliminary cleaning, the substantive generation and validation of reasoning logic were exclusively performed by the senior doctoral researchers to ensure rigor.

\textbf{Level 2} questions are designed to assess the model’s ability to comprehend policy intent, stakeholder interests, and institutional logic, following the 3I framework (Ideas, Interests, Institutions). These questions emphasize abstraction and interpretation rather than factual recall. \textbf{Level 3} questions focus on practical reasoning, including scenario-based decision-making, numerical calculations, and value-driven trade-offs, often grounded in real-world policy contexts.

All questions are constructed to be clear, faithful to source policies, and cognitively representative of their designated levels (see \autoref{fig:case-study} for some examples). To ensure quality and consistency, the resulting items undergo a human evaluation process (for details, see \autoref{sec:human_eval}).

\subsection{Details of \textit{PolicyMoE}}
\label{sec:training}

Our MoE architecture is initialized using \texttt{Qwen2.5-7B-Instruct} as the base model with bfloat16 precision. Expert modules are fine-tuned using LoRA with rank $r = 16$, scaling factor $\alpha = 32$, and dropout rate of 0.05, targeting the attention and MLP projection layers. Expert training is conducted for 3 epochs with a per-device batch size of 4 and gradient accumulation of 4 steps, resulting in an effective batch size of 16. The learning rate is set to $5 \times 10^{-5}$ with weight decay of 0.01. Router training follows for 2 epochs with batch size 8 and learning rate $1 \times 10^{-4}$, using a two-layer MLP architecture with LayerNorm and GELU activation. The dataset is split into 80\% training and 20\% testing, with maximum sequence lengths of 2048 tokens for expert training and 512 tokens for router training across the three domains: memory, comprehension, and application.

\begin{figure*}[t]
    \centering
    \includegraphics[width=1.0\linewidth]{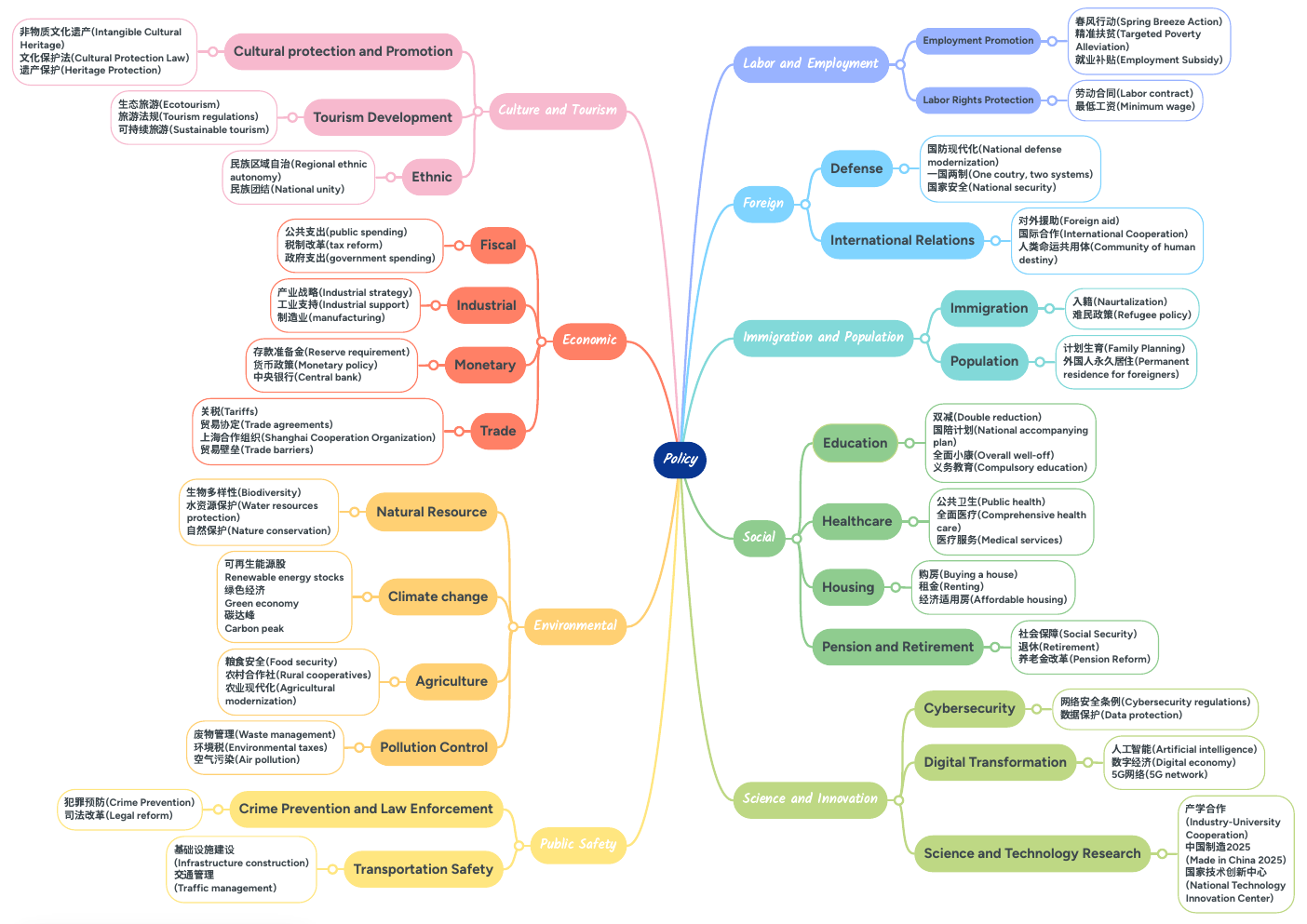}
    \caption{Categories of Chinese Policy Documents and Representative Keywords (Partial List).}
    \label{fig:keywords}
\end{figure*}

\section{Related Works}

\textbf{LLM Benchmarks in Social Science.} The rapid advancements in LLMs have enabled their application in social science research. The ability to efficiently comprehend long textual inputs and generate human-like responses makes them favorable for text-intensive tasks, like academic paper searching \cite{he2025pasallmagentcomprehensive,baek2025researchagentiterativeresearchidea,asai2024openscholarsynthesizingscientificliterature}, document analysis \citep{wang2023docllm} and legal research \citep{colombo2024saullm7bpioneeringlargelanguage} \citep{Savelka_2023}. Such ability underscores the need for tailored evaluation methods that measure their performance in these domains \cite{bao2026position}.  Knowledge-intensive benchmarks, such as MMLU \citep{wang2024mmlu} and TriviaQA \citep{joshi2017triviaqa}, test models on a wide range of subjects, from STEM fields to humanities, shedding light on the recalling and reasoning ability of LLMs to handle complex queries in real-world scenarios.

Domain-specific benchmarks now probe specialized knowledge: BioASQ \citep{tsatsaronis2015overview} tests biomedical QA, MedQA \citep{jin2021disease} evaluates clinical reasoning. Legal NLP has advanced through LawBench \citep{fei2024lawbench} for Chinese statutory analysis and LegalBench \citep{guha2023legalbench} for Anglo-American jurisprudence tasks. Recent works have begun to address LLM evaluation in the policy domain. For instance, PolicySimEval \citep{kang2025policysimeval} assesses policy outcomes through simulation, while UNBench \citep{liang2025benchmarking} evaluates performance on political science tasks from a UN perspective. These benchmarks differ from \textit{PolicyBench}, which focuses on the fine-grained comprehension of policy texts across broad domestic domains and governmental systems. In parallel, studies by Safaei et al. \citep{safaei2024end} and Karacapilidis et al. \citep{karacapilidis2024generative} explore the application of LLMs for policy generation and deliberation. While these efforts target the 'output' capabilities of LLMs, our work addresses a complementary and foundational gap: evaluating the model's 'input' capability to precisely understand policy language, a crucial prerequisite for any reliable downstream application.

\begin{figure}
    \begin{center}
         \includegraphics[width=1.0\linewidth]{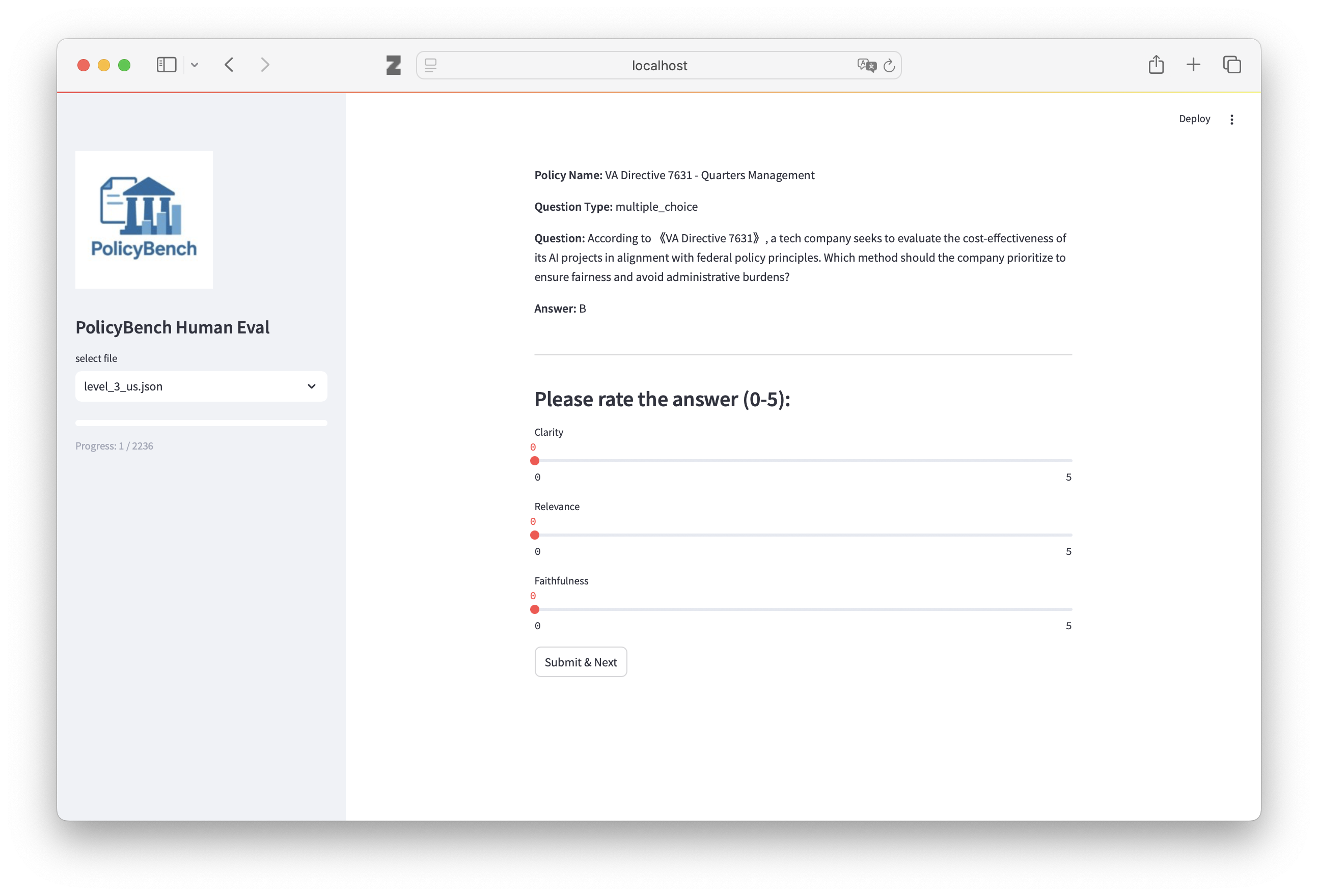}
    \end{center}
    \caption{Screenshot of human evaluation interface.}
    \label{fig:screenshot}
\end{figure}


\begin{table*}[htbp]
\centering
\caption{Sources for Policies and Supplementary Materials (By Language)}
\label{tab:data_sources}
\begin{tabular}{ll}
\toprule
\textbf{Language \& Content Type} & \textbf{Primary Collection Websites / Platforms} \\
\midrule
\rowcolor{gray!20}
\multicolumn{2}{l}{\textit{US Sources}} \\
Official Policies &  
\url{https://www.transportation.gov} \\ & 
\url{https://www.hhs.gov} \\ &
\url{https://www.va.gov} \\ &
\url{https://www.commerce.gov} \\ &
\url{https://www.usda.gov} \\ &
\url{https://www.energy.gov} \\ &
\url{https://www.doi.gov} \\ &
\url{https://www.ed.gov} \\ &
\url{https://www.treasury.gov} \\ &
\url{https://www.state.gov} \\ &
\url{https://www.dhs.gov} \\ &
\url{https://www.hud.gov} \\
Supplementary Materials &
\url{https://www.cnn.com/} \\ & 
\url{https:// www.foxnews.com} \\ & \url{https://www.reuters.com}\\
\midrule
\rowcolor{gray!20}
\multicolumn{2}{l}{\textit{Chinese Sources}} \\
Official Policies & \url{https://www.gov.cn/zhengce/zhengcewenjianku/} \\
Supplementary Materials & \url{https://www.xuexi.cn/} \\ & \url{http://www.people.com.cn/} \\
\bottomrule
\end{tabular}
\end{table*}

\begin{table*}[t]
\centering
\renewcommand{\arraystretch}{1.2}
\setlength{\tabcolsep}{6pt}
\caption{Overview of models evaluated or used as judges in this study.}
\label{tab:models}
\resizebox{\textwidth}{!}{
\begin{tabular}{lcccc}
\toprule
\textbf{Model} & \textbf{Developer} & \textbf{Open-source} & \textbf{Version} & \textbf{Role} \\
\midrule
GPT-4o & OpenAI & \textcolor{red}{\faTimesCircle} & / & Evaluated model \\
o4-mini &  & \textcolor{red}{\faTimesCircle} & 2025-04-16 & Evaluated model; Judge (Level 3) \\
\midrule
Gemini-2.5-Flash & Google DeepMind & \textcolor{red}{\faTimesCircle} & preview-04-17 & Evaluated model; Judge (Level 3) \\
Gemini-2.0-Flash &  & \textcolor{red}{\faTimesCircle} & / & Evaluated model \\
Gemma-3-27B &  & \textcolor{green!70!black}{\faCheckCircle} & 27B Instruct & Evaluated model \\
\midrule
Claude-3.7-Sonnet & Anthropic & \textcolor{red}{\faTimesCircle} & 20250219 & Evaluated model; Judge (Level 3) \\
Claude-3.5-Sonnet &  & \textcolor{red}{\faTimesCircle} & 20241022 & Evaluated model \\
\midrule
LLaMA-4 & Meta & \textcolor{green!70!black}{\faCheckCircle} & maverick-instruct & Evaluated model \\
\midrule
QwQ-32B & Alibaba (Qwen Team) & \textcolor{green!70!black}{\faCheckCircle} & / & Evaluated model \\
\midrule
DeepSeek-V3 & DeepSeek AI & \textcolor{green!70!black}{\faCheckCircle} & / & Evaluated model \\
DeepSeek-R1 &  & \textcolor{green!70!black}{\faCheckCircle} & / & Evaluated model; Judge (Level 3) \\
\bottomrule
\end{tabular}}
\end{table*}
\begin{table*}[t]
    \centering
    \small 
    \renewcommand\arraystretch{1.3} 
    \caption{The detailed introduction of 10 dimensions.}
    \label{tab:dimensiondefinition}
    \begin{tabularx}{\textwidth}{@{} l c >{\RaggedRight}X @{}}
        \toprule[1pt]
        \textbf{Dimension} & \textbf{ID} & \textbf{Definition} \\
        \midrule
        Article/Date Memorization & 1-1 & Tests memory of specific articles, dates, numbers, etc. in the policy. \\
        Terminology Recognition   & 1-2 & Tests ability to recognize and understand policy terminology. \\
        Organization Identification & 1-3 & Tests ability to identify organizations mentioned in the policy. \\
        Idea Understanding & 2-1 & Examines the ideological foundation and value orientation behind the policy. \\
        Interest Understanding & 2-2 & Assesses the identification and analysis of key stakeholders affected by or involved in the policy. \\
        Institution Understanding & 2-3 & Evaluates understanding of the formal and informal rules, organizations, and mechanisms shaping policy implementation. \\
        Policy-Based Numerical Reasoning & 3-1 & Perform simple mathematical reasoning or calculation based on the numerical provisions in the policy text. \\
        Scenario-Based Decision-Making & 3-2 & Based on specific scenarios, determine how the parties should make decisions or choose the most appropriate approach based on the policy. \\
        Procedural/Institutional Implementation & 3-3 & Examine the understanding and memory of the specific operational procedures, implementation steps, and institutional regulations in the policy. \\
        Policy Logic and Value Explanation & 3-4 & Focus on the background motivation, target value and logical structure of policy making. \\
        \bottomrule[1pt]
    \end{tabularx}
\end{table*}

\begin{figure*}[htbp]
    \centering
    \includegraphics[width=1.0\linewidth]{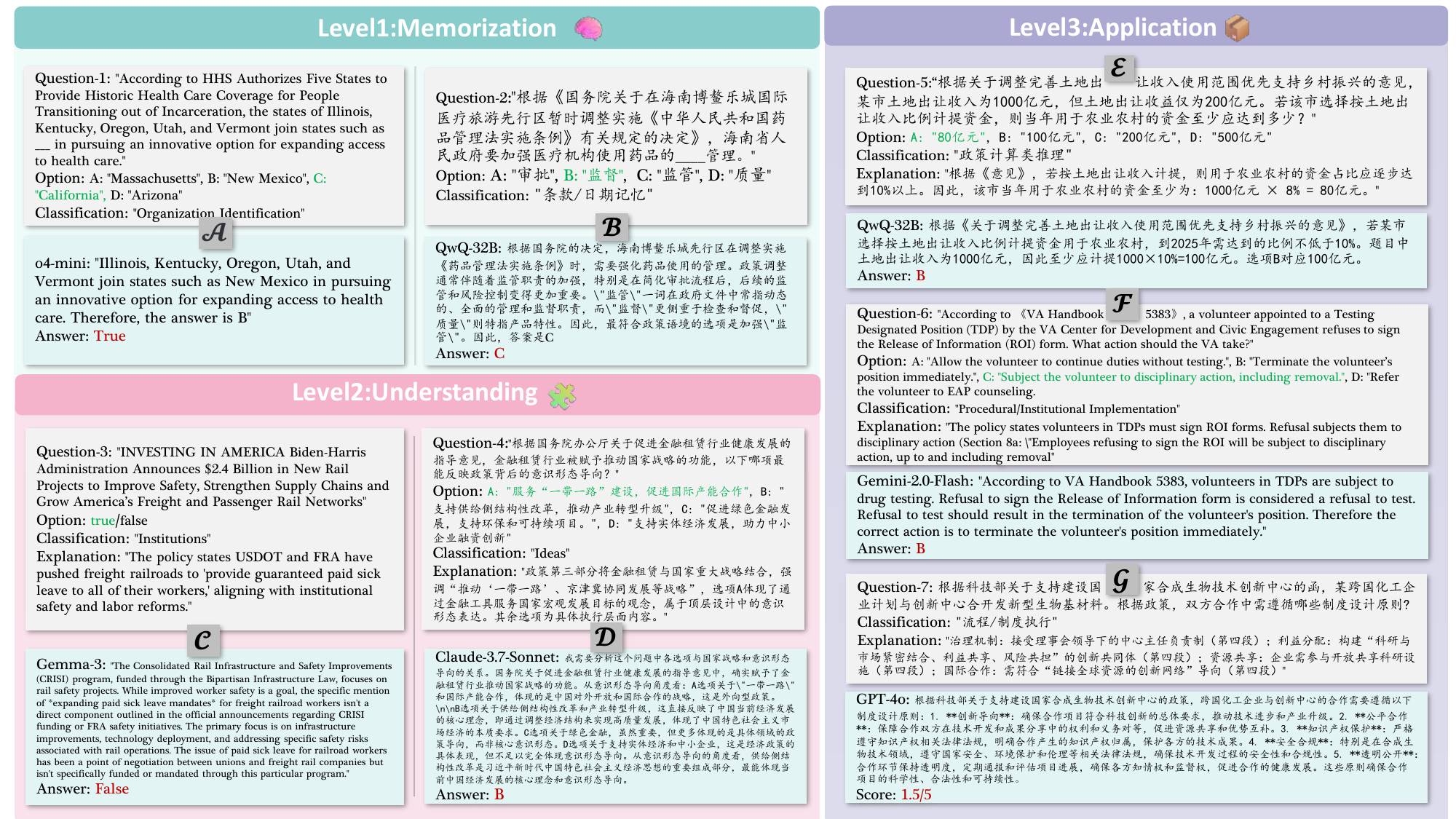}
    \vspace{-10pt}
    \caption{Representative error cases across three cognitive levels in \textit{PolicyBench}.}
    \label{fig:error-study}
    \vspace{-10pt}
\end{figure*}

\section{Error Study}
\label{sec:error_study}

To better understand the limitations of current LLMs on \textit{PolicyBench}, we conducted a qualitative error analysis by sampling representative failure cases from each level, as illustrated in \autoref{fig:error-study}.

\noindent\textbf{\ul{\textit{Level 1 (Memorization): Factual Confusions.}}} The most common errors involve incorrect recall or misidentification, often triggered by the presence of distractors with strong semantic similarity. For example, when asked to identify US states that previously adopted a healthcare policy, the model incorrectly selected “New Mexico” instead of “California”—likely due to co-occurrence bias in training data. In Chinese policy questions, LLMs sometimes confuse regulatory terms with subtle distinctions (e.g., “Jiān Dū” vs. “Jiān Guǎn”), indicating limited sensitivity to nuanced legal language.

\noindent\textbf{\ul{\textit{Level 2 (Understanding): Misinterpretations.}}}
Errors at this stage largely stem from misreading policy intent, ideological framing, or institutional logic. Models tend to misread underlying motivations or over-rely on surface cues. For instance, a model incorrectly identified “Supply-side structural reform” as the core ideological signal of a financial leasing policy, despite explicit references to national strategies like the Belt and Road Initiative. Similarly, when analyzing US labor protection clauses, the model missed the correct interpretation of sick leave provisions, favoring superficial summaries over deeper institutional mandates described in the text.

\noindent\textbf{\ul{\textit{Level 3 (Application): Reasoning and Procedural Failures.}}} 
Models frequently struggled with quantitative reasoning, procedural interpretation, and hallucinated conclusions. For example, a model failed to compute the required rural funding quota from land sale revenue, despite having all necessary information. In another case, a model prematurely recommended immediate termination for a volunteer’s non-compliance, overlooking the stepwise disciplinary procedures mandated by policy. For open-ended tasks, models sometimes hallucinate plausible-sounding but unsupported principles, rather than extracting the specific cooperation principles explicitly mentioned in the document.

Overall, these error patterns reveal that LLMs struggle across recall, understanding, and real-world reasoning, especially with legal nuance and institutional complexity. Addressing these issues may require targeted supervision or retrieval-augmented approaches.

\section{Data Quality \& Human Evaluation}
\label{sec:human_eval}

\subsection{Annotation Team \& Methodology}
To ensure the rigorousness of \textit{PolicyBench}, our annotation team was structured hierarchically. The core annotation, quality control, and validation phases were led by \textbf{five senior Ph.D. candidates} specializing in Public Policy, Law, and Computational Social Science. Three undergraduate assistants provided support for preliminary data formatting and cleaning but did not perform substantive labeling or validation tasks.

\subsection{General Quality Evaluation}
We conducted a comprehensive human evaluation between March 2025 and April 2025. Each generated question was independently evaluated along three key dimensions on a 1--5 Likert scale. We sampled \textbf{1,500} (Level 1), \textbf{1,000} (Level 2), and \textbf{500} (Level 3) questions balanced across languages.

As shown in \autoref{tab:human_eval}, the dataset exhibits consistently high quality, with average scores exceeding \textbf{96\%} across all dimensions.

\begin{table*}[t]
\centering
\caption{Human evaluation results across levels and regions.}
\label{tab:human_eval}
\setlength{\tabcolsep}{12pt}
\begin{tabular}{llcccc}
\toprule[1pt]
\textbf{Level} & \textbf{Region} & \textbf{Clarity} & \textbf{Relevance} & \textbf{Faithfulness} & \textbf{Avg.} \\
\midrule
\multirow{2}{*}{\textbf{Level 1}} & \textbf{CN}   & 99.20\% & 98.01\% & 98.78\% & 98.66\% \\
                                    & \textbf{US} & 99.15\% & 98.25\% & 97.94\% & 98.45\% \\
\midrule
\multirow{2}{*}{\textbf{Level 2}} & \textbf{CN}   & 98.86\% & 98.00\% & 98.12\% & 98.32\% \\
                                    & \textbf{US} & 98.76\% & 97.60\% & 97.22\% & 97.86\% \\
\midrule
\multirow{2}{*}{\textbf{Level 3}} & \textbf{CN}   & 99.16\% & 96.02\% & 96.14\% & 97.11\% \\
                                    & \textbf{US} & 97.66\% & 95.64\% & 96.32\% & 96.54\% \\
\bottomrule[1pt]
\end{tabular}
\vspace{-10pt}
\end{table*}

\subsection{Inter-Annotator Agreement (IAA)}
To validate the reliability of the human evaluation, we performed a double-blind annotation on a random subset of 600 items (20\% of the evaluation set). We utilized \textit{\textbf{Cohen's Kappa ($\kappa$)}} to measure agreement beyond chance, calculated as:
\begin{equation}
    \kappa = \frac{P_o - P_e}{1 - P_e}
\end{equation}
where $P_o$ is the observed agreement and $P_e$ is the expected agreement by chance.

\textbf{Calculation Method:}
\begin{itemize}[noitemsep,leftmargin=*]
    \item For \textbf{Bloom-level classification} (Nominal), we calculated standard unweighted $\kappa$.
    \item For \textbf{Quality dimensions} (Ordinal 1--5), we \textbf{binarized} the scores into ``Accept'' (Score $\ge 4$) and ``Reject'' (Score $\le 3$) to strictly measure the consistency of inclusion criteria.
\end{itemize}

As shown in \autoref{tab:iaa_stats}, we achieved strong agreement ($\kappa > 0.80$) across all dimensions. Disagreements were resolved via a two-stage adjudication process involving a senior expert.

\begin{table}[h]
\centering
\caption{Inter-Annotator Agreement statistics. Quality scores were binarized for $\kappa$ calculation.}
\label{tab:iaa_stats}
\resizebox{\linewidth}{!}{
\begin{tabular}{lcc}
\toprule
\textbf{Annotation Dimension} & \textbf{Kappa ($\kappa$)} & \textbf{Agreement Level} \\
\midrule
Bloom Classification (L1/L2/L3) & 0.856 & Strong \\
Clarity (Pass/Fail) & 0.841 & Strong \\
Relevance (Pass/Fail) & 0.877 & Strong \\
Faithfulness (Pass/Fail) & 0.827 & Strong \\
\bottomrule
\end{tabular}
}
\vspace{-5pt}
\end{table}

\subsection{Expert Validation Study}
To further verify the dataset's validity against a professional standard, we conducted an additional expert validation study with four external experts (\textit{\textbf{2 Ph.D. candidates in Public Policy, 1 Professor in Computational Social Science, and 1 Senior Sociology Researcher}}). They evaluated a stratified sample of 120 questions.

\textbf{1. Data Validity:}
Experts assessed the correctness of the Gold Answers and the appropriateness of the content. Agreement for Bloom-level labeling was calculated using \textit{\textbf{Krippendorff’s $\alpha$}}, which is robust for multiple raters:
\begin{equation}
    \alpha = 1 - \frac{D_o}{D_e}
\end{equation}
where $D_o$ is the observed disagreement and $D_e$ is the expected disagreement by chance. The results in \autoref{tab:expert_validity} confirm the dataset's high quality.

\begin{table}[h]
\centering
\caption{Expert Validation Results ($N=120$).}
\label{tab:expert_validity}
\resizebox{\linewidth}{!}{
\begin{tabular}{lc}
\toprule
\textbf{Metric} & \textbf{Score} \\
\midrule
\textbf{Correctness Verification} & 96.1\% \\
\textbf{Content Validity} (Rated ``Appropriate'') & 94.5\% \\
\textbf{Bloom-Label Agreement} (Krippendorff’s $\alpha$) & 0.86 \\
\bottomrule
\end{tabular}
}
\vspace{-5pt}
\end{table}

\textbf{2. Expert Performance Baseline:}
To establish a human ceiling, the experts answered the questions under an \textbf{open-book setting}, simulating realistic policy analysis workflows. As shown in \autoref{tab:expert_performance}, experts significantly outperform models in deeper understanding tasks (L2/L3), validating the benchmark's difficulty gradient.

\begin{table}[h!]
\centering
\caption{Human Expert Performance (Open-Book) vs. Models.}
\label{tab:expert_performance}
\resizebox{\linewidth}{!}{
\begin{tabular}{lcc}
\toprule
\textbf{Level} & \textbf{Expert Accuracy} & \textbf{Comparison vs. SOTA LLM} \\
\midrule
Level 1 (Memorization) & 82.3\% & Comparable \\
Level 2 (Understanding) & \textbf{88.4\%} & \textbf{Significant Gap} \\
Level 3 (Application) & \textbf{90.1\%} & \textbf{Significant Gap} \\
\bottomrule
\end{tabular}
}
\end{table}

\section{Formal Definition of the Policy Tasks}
\label{app:task_definition}

To address reviewer concerns regarding the lack of a formal task definition, we provide here a unified formulation of the \emph{Policy Comprehension Task}, followed by fine-grained instantiations corresponding to the three cognitive levels and ten sub-tasks, details in \autoref{tab:dimensiondefinition}.

\subsection{General Formulation}

At a high level, we model policy comprehension as a conditional question-answering problem grounded in policy documents. Let $C$ denote a policy context, which may consist of a full policy document, a section, or a clause describing rules, actors, and institutional mechanisms. Let $Q_k$ denote a query designed to probe a specific cognitive level $k \in \{1,2,3\}$, corresponding to memorization, understanding, and application, respectively.

Given $(C, Q_k)$, a language model $f_\theta$ is expected to produce an answer $\hat{A}$ that is consistent with the policy content and satisfies the cognitive requirement implied by $k$:
\begin{equation}
\hat{A} = \arg\max_{A \in \mathcal{A}} P(A \mid C, Q_k; \theta),
\end{equation}
where $\mathcal{A}$ denotes the answer space, which can be either a finite set of options (for multiple-choice questions) or free-form text (for open-ended questions).

\subsection{Task Instantiation by Cognitive Level}

We instantiate the policy comprehension task into \textbf{10 concrete sub-tasks}, organized into three cognitive levels.

\paragraph{Level 1: Memorization (Factual Retrieval)}
\emph{Objective: Retrieve explicit facts or entities directly stated in the policy text $C$.}

\subparagraph{\textit{1-1 Article / Date Memorization}}
Recall specific temporal markers or citation identifiers.
\begin{quote}
\emph{Example:} ``According to the \textit{U.S.-Philippines Civil Nuclear Cooperation Agreement}, the Agreement entered into force on [Mask].''\\
\emph{Answer:} July~2.
\end{quote}

\subparagraph{\textit{1-2 Terminology Recognition}}
Identify the definition of a domain-specific term as stated in the text.
\begin{quote}
\emph{Example:} ``Civil nuclear cooperation agreements provide a legal framework for exports of [Mask].''\\
\emph{Answer:} material, equipment, and components.
\end{quote}

\subparagraph{\textit{1-3 Organization Identification}}
Identify organizational hierarchy or institutional affiliation.
\begin{quote}
\emph{Example:} ``According to the \textit{National Behavioral Health Workforce Career Navigator}, SAMHSA is an agency within [Mask].''\\
\emph{Answer:} U.S. Department of Health and Human Services (HHS).
\end{quote}

\paragraph{Level 2: Understanding (Conceptual Interpretation)}
\emph{Objective: Map explicit policy text to implicit concepts under the ``3I'' framework (Ideas, Interests, Institutions).}

\subparagraph{\textit{2-1 Idea Understanding}}
Infer the underlying goal, ideology, or strategic intent.
\begin{quote}
\emph{Example:} ``The \textit{Big Data Project (BDP)} aims to democratize access to environmental data primarily through collaboration with whom?''\\
\emph{Answer:} Cloud Service Providers.
\end{quote}

\subparagraph{\textit{2-2 Interest Understanding}}
Identify stakeholders and their eligibility, benefits, or losses.
\begin{quote}
\emph{Example:} ``Are Tribal entities eligible to apply for 2501 Program grants according to the USDA announcement?''\\
\emph{Answer:} True.
\end{quote}

\subparagraph{\textit{2-3 Institution Understanding}}
Comprehend rules, categories, or allocation mechanisms.
\begin{quote}
\emph{Example:} ``Which category under the \textit{Clean Energy} program allocates capacity to projects ensuring 50\% of benefits go to low-income households?''\\
\emph{Answer:} Category~4.
\end{quote}

\paragraph{Level 3: Application (Scenario Reasoning)}
\emph{Objective: Apply policy rules to a novel or hypothetical scenario.}

\subparagraph{\textit{3-1 Policy-Based Numerical Reasoning}}
Execute numerical calculations derived from policy formulas.
\begin{quote}
\emph{Example:} ``If a \$2 million FEMA project has an 80\% federal cost share and a 1.91\% cost increase, how much does the local government pay?''\\
\emph{Answer:} \$7{,}640.
\end{quote}

\subparagraph{\textit{3-2 Scenario-Based Decision Making}}
Determine the compliant action in a simulated situation.
\begin{quote}
\emph{Example:} ``If the Colonial Pipeline were subject to the new safety rule during a leak, which action would be required?''\\
\emph{Answer:} Stage response personnel in predefined zones.
\end{quote}

\subparagraph{\textit{3-3 Procedural / Institutional Implementation}}
Validate procedural conditions or sequences.
\begin{quote}
\emph{Example:} ``What is a necessary condition for the transfer of nuclear reactors under the U.S.-Thailand 123 Agreement?''\\
\emph{Answer:} Commitment to nonproliferation standards.
\end{quote}

\subparagraph{\textit{3-4 Policy Logic and Value Explanation}}
Explain conflicts or trade-offs between policy objectives.
\begin{quote}
\emph{Example:} ``Scott Turner’s goal to reduce reliance on government aid conflicts with which aspect of DC’s housing strategy?''\\
\emph{Answer:} Funding for emergency rental assistance.
\end{quote}

This structured definition clarifies both the scope and the granularity of policy comprehension capabilities evaluated by \textit{PolicyBench}.

\section{Multi-Examiner Bias Analysis}
\label{sec:bias_analysis}

To address critical concerns regarding \textit{"Model-Speak"} (stylistic cues) and \textit{"Familiarity Bias"} (models favoring their own generation patterns), we conducted a comprehensive \textbf{Multi-Examiner Sensitivity Analysis}. This study empirically validates that our heterogeneous examiner pool serves as a necessary safeguard against evaluation artifacts.

\subsection{Experimental Setup}

We designed a controlled experiment to decouple the "Examiner" (Question Generator) from the "Examinee" (Evaluated Model).

\noindent\textbf{1. The Examiner Pool:}
We utilized three distinct top-tier models to generate questions and distractors, ensuring coverage across different model families:
\begin{itemize}[noitemsep,leftmargin=*]
    \item \textbf{GPT Family:} \texttt{GPT-4o} (OpenAI)
    \item \textbf{Claude Family:} \texttt{Claude-4-Sonnet} (Anthropic)
    \item \textbf{Qwen Family:} \texttt{Qwen-3} (Alibaba Cloud)
\end{itemize}

\noindent\textbf{2. The Examinee Pool:}
We evaluated 7 models, including the three generator families and external open-weights models, to observe cross-family behaviors:
\begin{itemize}[noitemsep,leftmargin=*]
    \item \textit{Closed-Source:} \texttt{GPT-4o, GPT-4o-mini, Claude-4-Sonnet, Claude-4-Haiku}.
    \item \textit{Open-Weights:} \texttt{Qwen-3, Qwen-2.5, and Llama-4}.
\end{itemize}

\noindent\textbf{3. Evaluation Conditions:}
We tested these models across three distinct settings:
\begin{itemize}[noitemsep,leftmargin=*]
    \item \textbf{Baseline (Consensus):} Questions generated by the full pool. This is the standard \textit{PolicyBench} setting.
    \item \textbf{Single-Examiner:} Questions generated exclusively by one examiner (e.g., \textit{GPT-Only}).
    \item \textbf{LOEO (Leave-One-Examiner-Out):} Questions generated by the remaining two examiners (e.g., \textit{Wo-GPT}).
\end{itemize}

\subsection{Full Leaderboard Sensitivity Results}

\autoref{tab:full_sensitivity} presents the complete performance matrix. The results demonstrate significant score variations under single-examiner conditions, confirming the necessity of our Consensus Baseline.

\begin{table*}[t]
\centering
\caption{Complete Multi-Examiner Sensitivity Analysis. Scores represent accuracy (\%). \textbf{"Baseline"} denotes the standard PolicyBench setting (3-Examiner Consensus). \textbf{"Wo-X"} denotes the Leave-One-Examiner-Out setting. The data reveals that relying on a single examiner (e.g., GPT-Only or Claude-Only) leads to drastic score fluctuations compared to the stable Baseline.}
\label{tab:full_sensitivity}
\resizebox{\textwidth}{!}{
\begin{tabular}{l|c|ccc|ccc|c}
\toprule
\multirow{2}{*}{\textbf{Model}} & \textbf{Baseline} & \multicolumn{3}{c|}{\textbf{Single-Examiner Setting}} & \multicolumn{3}{c|}{\textbf{LOEO Setting}} & \multirow{2}{*}{\textbf{Avg.}} \\
 & (Consensus) & \textbf{Claude-Only} & \textbf{GPT-Only} & \textbf{Qwen-Only} & \textbf{Wo-Claude} & \textbf{Wo-GPT} & \textbf{Wo-Qwen} &  \\
\midrule
\textbf{Qwen-3} & 89.0 & 86.0 & 88.0 & \textbf{92.0} & \textbf{92.0} & \textbf{90.0} & 88.0 & 89.29 \\
\textbf{Llama-4} & 82.0 & 84.0 & 88.0 & 89.0 & 88.0 & 87.0 & 85.0 & 86.14 \\
\textbf{Qwen-2.5} & 83.0 & 81.0 & 85.0 & 90.0 & 83.0 & 86.0 & \textbf{87.0} & 85.00 \\
\textbf{GPT-4o-mini} & 66.0 & 59.0 & 74.0 & 85.0 & 74.0 & 71.0 & 67.0 & 82.67 \\
\textbf{GPT-4o} & 75.0 & 71.0 & \textbf{82.0} & 85.0 & 85.0 & 83.0 & 78.0 & 79.86 \\
\textbf{Claude-4-Sonnet} & 84.0 & 69.0 & 49.0 & 89.0 & 71.0 & 52.0 & 85.0 & 71.29 \\
\textbf{Claude-4-Haiku} & 60.0 & 47.0 & 81.0 & 86.0 & 59.0 & 48.0 & 75.0 & 91.20 \\
\bottomrule
\end{tabular}
}
\vspace{-5pt}
\end{table*}

\subsection{Analysis of Biases}

\noindent\textbf{1. Self-Scoring Bias (Familiarity).}
Models consistently perform differently on questions they generated themselves, creating a "Familiarity Bonus" or "Penalty". As shown in \autoref{tab:self_bias}, relying on a single generator creates severe distortions:

\begin{table*}[h]
\centering
\caption{Analysis of Self-Scoring Bias. The results show that single-examiner benchmarks suffer from extreme variance (from +7 inflation to -15 deflation).}
\label{tab:self_bias}
\resizebox{\linewidth}{!}{
\begin{tabular}{lcccc}
\toprule
\textbf{Model} & \textbf{Baseline Score} & \textbf{Own-Gen Score} & \textbf{$\Delta$} & \textbf{Bias Type} \\
\midrule
\textbf{GPT-4o} & 75.0\% & 82.0\% & \textcolor{red}{\textbf{+7.0\%}} & \textit{Self-Leniency / Pattern Matching} \\
\textbf{Qwen-3} & 89.0\% & 92.0\% & \textcolor{red}{+3.0\%} & \textit{Moderate Inflation} \\
\textbf{Claude-4-Sonnet} & 84.0\% & 69.0\% & \textcolor{blue}{\textbf{-15.0\%}} & \textit{Self-Strictness / Hyper-Critical} \\
\bottomrule
\end{tabular}
}
\end{table*}

\noindent\textbf{Insight:} The data reveals divergent biases. GPT-4o benefits from "Familiarity Bias" (+7\%), likely exploiting its own stylistic patterns. Conversely, Claude exhibits "Self-Strictness" (-15\%), penalizing its own generation logic. \textbf{PolicyBench's consensus approach effectively averages out these extremes}, anchoring scores to a neutral ground truth.

\noindent\textbf{2. Mitigating Model-Speak: Leaderboard Stability.}
To determine if "Model-Speak" (stylistic tells) compromises the validity of the rankings, we analyzed the \textbf{Spearman Rank Correlation ($\rho$)} \cite{zar2005spearman} between the Baseline leaderboard and other conditions.

\begin{table}[h]
\centering
\caption{Leaderboard Stability Analysis. High correlation in LOEO conditions confirms the robustness of the ranking system, while single-examiner rankings (especially GPT-Only) are highly unstable.}
\label{tab:rank_stability}
\resizebox{\linewidth}{!}{
\begin{tabular}{lccc}
\toprule
\textbf{Comparison Pair} & \textbf{Spearman $\rho$} & \textbf{Kendall $\tau$} & \textbf{Interpretation} \\
\midrule
Baseline vs. \textbf{Wo-Qwen} & \textbf{0.901} & 0.781 & \textit{Highly Stable} \\
Baseline vs. \textbf{Wo-GPT} & 0.607 & 0.524 & \textit{Moderately Stable} \\
Baseline vs. \textbf{GPT-Only} & \textbf{0.342} & 0.293 & \textit{\textbf{Unstable / Biased}} \\
\bottomrule
\end{tabular}
}
\end{table}

\noindent\textbf{3. External Model Robustness.}
For models outside the generator pool, like \textbf{Llama-4}, the Baseline provides the most stable evaluation. Llama-4's score varies from 82.0\% to 89.0\% across single examiners. The Baseline (82.0\%) successfully anchors it to a consensus difficulty, filtering out examiner-specific noise.

\textbf{Conclusion:} The low correlation of the GPT-Only set ($\rho=0.34$) proves that single-source benchmarks are fundamentally biased. The multi-examiner design in \textit{PolicyBench} is a necessary mechanism to ensure that high performance reflects genuine \textit{Policy Comprehension} rather than \textit{Stylistic Alignment}.

\section{LLM-as-a-Judge Validation}
\label{sec:judge_validation}

To ensure the reliability and validity of our automated evaluation pipeline, we conducted a two-fold validation process: assessing \textit{stability} (consistency across runs) and \textit{alignment} (accuracy against human experts).

\subsection{Evaluation Stability Analysis}
\label{sec:stable}

\noindent\textbf{Evaluation Stability.} To rigorously assess the stability of our LLM-as-a-Judge pipeline and address concerns about the stochastic nature of model outputs, we conducted a multi-run, multi-judge evaluation analysis. We randomly sampled 10 question-answer pairs from the Level 3 open-ended test set. The scoring for each sample was performed as follows:

\begin{itemize}[noitemsep,leftmargin=*]
\item \textbf{Initial Scoring:} For each evaluation run, two distinct LLM judges were sampled from our pool to score the response on a 0--5 scale (in 0.5 increments). The score for the run was the average of these two ratings.
\item \textbf{Discrepancy Resolution:} If the scores from the two initial judges differed by more than 1.0 point, a third tie-breaker judge was invoked to provide an additional score. In such cases, the final score for the run was the average of all three judges' ratings. This protocol ensures robustness against outlier judgments.
\item \textbf{Repetition:} This entire scoring process was conducted three independent times for each of the 10 cases.
\end{itemize}

We then calculated the mean and standard deviation of the three final scores for each case. The results, summarized in \autoref{tab:stability_summary}, demonstrate high consistency, with standard deviations remaining exceptionally low. This indicates that our multi-judge protocol effectively mitigates run-to-run variance and produces reliable evaluations. 

\begin{table*}[h]
\centering
\caption{Stability analysis of the final LLM-as-a-Judge scores. The low standard deviation across three independent runs for each case demonstrates the reliability of our multi-judge evaluation protocol.}
\label{tab:stability_summary}
\resizebox{\linewidth}{!}{
\begin{tabular}{@{}lccccc@{}}
\toprule
\textbf{Case ID} & \textbf{Final Score (Run 1)} & \textbf{Final Score (Run 2)} & \textbf{Final Score (Run 3)} & \textbf{Mean} & \textbf{Std. Dev.} \\ \midrule
Case 1        & 4.00                         & 4.50                         & 4.00                         & 4.17          & 0.29              \\
Case 2        & 4.00                         & 4.00                         & 4.25                         & 4.08          & 0.12              \\
Case 3        & 3.00                         & 3.25                         & 3.00                         & 3.08          & 0.12              \\
Case 4        & 4.33                         & 4.50                         & 4.25                         & 4.36          & 0.10              \\
Case 5        & 2.25                         & 2.25                         & 2.50                         & 2.33          & 0.12              \\
Case 6        & 5.00                         & 5.00                         & 5.00                         & 5.00          & 0.00              \\
Case 7        & 3.67                         & 4.00                         & 3.50                         & 3.72          & 0.25              \\
Case 8        & 1.50                         & 1.25                         & 1.00                         & 1.25          & 0.20              \\
Case 9        & 4.00                         & 3.75                         & 4.00                         & 3.92          & 0.12              \\
Case 10       & 4.67                         & 5.00                         & 4.50                         & 4.72          & 0.25              \\ \bottomrule
\end{tabular}
}
\end{table*}

\subsection{Human-Model Alignment}
\label{sec:alignment}

Stability is a necessary but insufficient condition for validity. To demonstrate that our LLM-as-a-Judge pipeline accurately reflects expert judgment, we conducted a \textbf{Human-Model Alignment Study}.

\noindent\textbf{Experimental Setup.} We sampled a subset of open-ended responses from Level 3 tasks. These responses were independently scored by our LLM Judge pipeline and by senior human experts (Ph.D. candidates in Public Policy) using the exact same rubric.

\noindent\textbf{Quantitative Results.} As shown in \autoref{tab:alignment_stats}, the LLM Judge demonstrates strong alignment with human experts across three key metrics:
\begin{itemize}[noitemsep,leftmargin=*]
    \item \textbf{Pearson Correlation ($r=0.87$):} Indicates a strong positive linear relationship between model and human scores.
    \item \textbf{Mean Absolute Error (MAE $= 0.42$):} On average, the model's score deviates from the human score by less than 0.5 points (the smallest scoring increment).
    \item \textbf{Agreement Rate ($94\%$):} In 94\% of cases, the model's score fell within an acceptable margin ($\le 1.0$ point) of the expert score.
\end{itemize}

\begin{table}[t]
\centering
\caption{Human--Model alignment statistics. High correlation, low error, and strong agreement indicate that the LLM judge closely matches expert evaluation.}
\label{tab:alignment_stats}
\setlength{\tabcolsep}{6pt}
\begin{tabular}{lc}
\toprule
\textbf{Metric} & \textbf{Value} \\
\midrule
Pearson Correlation ($r$) & \textbf{0.87} \\
Mean Absolute Error (MAE) & \textbf{0.42} \\
Agreement Rate            & \textbf{94\%} \\
\bottomrule
\end{tabular}
\end{table}

\noindent\textbf{Mitigating Subjectivity via Rubric-Based Scoring.}
The high alignment is primarily attributed to our \textbf{Point-Based Rubric} design (detailed in Appendix G). Unlike generic "quality" assessments which can be subjective, our prompt explicitly directs the model to verify the presence of specific \textit{Key Points} derived from the reference answer. The model assigns partial credit based on these matched points rather than an abstract "feeling" of quality. This structured approach significantly reduces hallucination and subjectivity, ensuring the judge acts as an objective verifier.

\section{Correlation with External Benchmarks}
\label{sec:external_benchmarks}

To demonstrate that \textit{PolicyBench} evaluates a distinct capability orthogonal to general reasoning or pure legal knowledge, we analyzed the performance correlation between \textit{PolicyBench} and two representative benchmarks: \textbf{MMLU-Pro} \cite{wang2024mmlu} (General Reasoning) and \textbf{LegalBench} \cite{guha2023legalbench} (Legal Reasoning).

\begin{table*}[h]
\centering
\caption{Performance comparison and correlation analysis across benchmarks.}
\label{tab:external_correlation}
\resizebox{1\linewidth}{!}{
\begin{tabular}{lcccc}
\toprule
\textbf{Model} & \textbf{MMLU-Pro} & \textbf{LegalBench} & \textbf{\textit{PolicyBench} (Avg)} & \textbf{\textit{PolicyBench} (L2)} \\
\midrule
DeepSeek-V3 & \textbf{81.9\%} & 80.1\% & 59.10\% & 57.68\% \\
GPT-4o & 80.3\% & 79.8\% & 59.47\% & 56.08\% \\
Claude-3.7-Sonnet & 80.3\% & 78.1\% & \textbf{64.13\%} & 58.48\% \\
Gemini-2.5-Flash & 77.9\% & \textbf{81.7\%} & 63.82\% & \textbf{64.06\%} \\
\midrule
\textbf{Pearson Correlation ($r$)} & \textbf{-0.69} & \textbf{-0.07} & \textbf{1.00} & -- \\
\bottomrule
\end{tabular}
}
\end{table*}

\textbf{Key Findings:}
\begin{itemize}[noitemsep,leftmargin=0pt]
    \item \textbf{Negative Correlation with General Reasoning ($r \approx -0.69$):} Surprisingly, models with top-tier general reasoning (e.g., \texttt{DeepSeek-V3}) do not necessarily excel in policy comprehension. This suggests that policy analysis involves specific logic (e.g., institutional constraints) that general benchmarks fail to capture.
    \item \textbf{No Correlation with Legal Reasoning ($r \approx -0.07$):} The near-zero correlation with LegalBench indicates that \textit{understanding policy} (Ideas, Interests) is fundamentally different from \textit{applying law} (Statutes). \textit{PolicyBench} fills this critical gap in the evaluation landscape.
\end{itemize}

\section{Comparing \textit{PolicyMoE} with a Standard LoRA}
\label{sec:ablation}

To validate the architectural contribution of \textit{PolicyMoE} and demonstrate its superiority over standard parameter-efficient fine-tuning, we conducted a controlled ablation study. We compared our \textit{PolicyMoE} architecture against a well-tuned \textbf{Standard LoRA} baseline.

\subsection{Experimental Setup}
To ensure a fair comparison, both models were trained on the same stratified subset of the PolicyBench training data using \texttt{Qwen2.5-7B-Instruct} as the backbone.
\begin{itemize}[noitemsep,leftmargin=*]
    \item \textbf{Standard LoRA:} Fine-tuned using a single LoRA adapter (Rank=16, Alpha=32) applied to all target modules, treating all levels of tasks as a unified objective.
    \item \textbf{\textit{PolicyMoE} (Ours):} Fine-tuned using our routing architecture with three specialized experts (Memory, Understanding, Application), utilizing the same data and hyperparameters.
\end{itemize}

\subsection{Results \& Analysis}
As shown in \autoref{tab:ablation}, \textit{PolicyMoE} outperforms the Standard LoRA baseline across all metrics, with an average accuracy improvement of \textbf{+2.68\%}.

\begin{table*}[h]
\centering
\caption{Performance comparison between the base model, Standard LoRA, and \textit{PolicyMoE} on a controlled training subset. \textit{PolicyMoE} achieves larger gains on Memorization (L1) and Application (L3) tasks.}
\label{tab:ablation}
\resizebox{\linewidth}{!}{
\begin{tabular}{lcccc}
\toprule
\textbf{Model} & \textbf{Level 1} & \textbf{Level 2} & \textbf{Level 3} & \textbf{Average} \\
 & (Memorization) & (Understanding) & (Application) &  \\
\midrule
Base (Qwen2.5-7B) & 30.10\% & 44.00\% & 55.69\% & 43.26\% \\
Standard LoRA & 34.82\% & 44.51\% & 59.45\% & 46.26\% \\
\textbf{\textit{PolicyMoE} (Ours)} & \textbf{38.63\%} & \textbf{44.90\%} & \textbf{63.30\%} & \textbf{48.94\%} \\
\midrule
\textit{$\Delta$ (MoE vs. LoRA)} & \textit{\textbf{+3.81\%}} & \textit{+0.39\%} & \textit{\textbf{+3.85\%}} & \textit{\textbf{+2.68\%}} \\
\bottomrule
\end{tabular}
}
\end{table*}


\noindent\textbf{Mitigating Task Interference.}
These results suggest that the MoE architecture helps alleviate \textit{task interference} arising from heterogeneous cognitive objectives.
\begin{itemize}[noitemsep,leftmargin=*]
    \item \textbf{Decoupling Memorization and Reasoning:} Standard LoRA must encode both exact factual recall (Level~1) and flexible scenario reasoning (Level~3) within a single low-rank adaptation, which can lead to competing gradient signals.
    \item \textbf{Improved Performance on Level 1:} By routing Level~1 queries to a dedicated \textbf{Memory Expert}, \textit{PolicyMoE} achieves a \textbf{+3.81\%} improvement over Standard LoRA, suggesting that specialized experts help preserve precise factual representations.
    \item \textbf{Implication:} These findings indicate that \textit{PolicyMoE} provides a structurally meaningful extension over standard LoRA when adapting LLMs to multi-level policy comprehension tasks.
\end{itemize}

\section{Selected Policy Samples}

\begin{figure*}
    \begin{center}
         \includegraphics[width=1.0\linewidth]{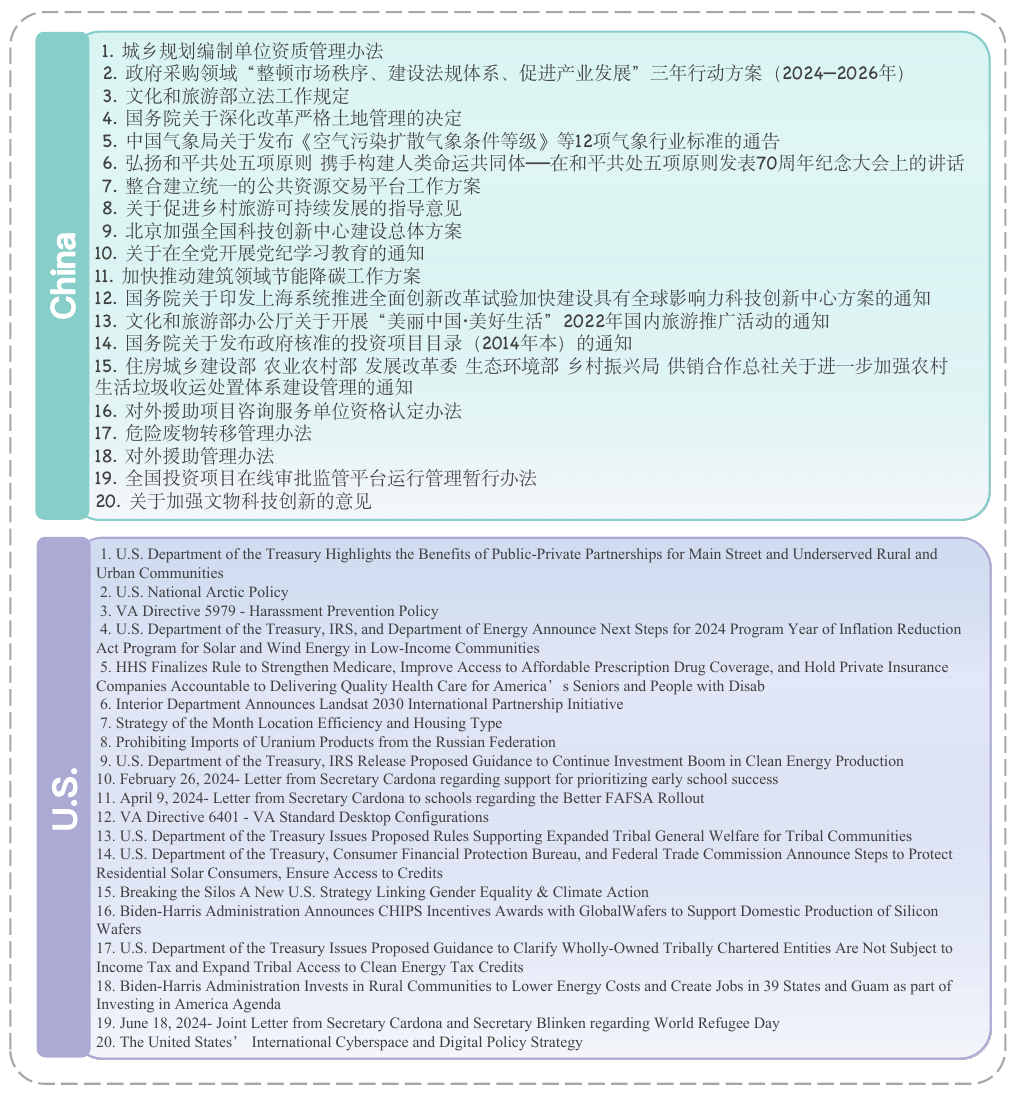}
    \end{center}
    \caption{Collected Policies (Part).}
    \label{fig:policy}
\end{figure*}

This section showcases examples of the policy titles from our dataset, from both China and the United States. A partial list is provided in \autoref{fig:policy}.

\section{Prompt Template}
This section shows the 3 key prompts used for data curation. \hyperref[prompt:1]{Prompt 1} converts clean policy text into cloze-style questions. \hyperref[prompt:2]{Prompt 2} instructs LLMs to generate incorrect answers, which serve as the distractors for multiple-choice questions. \hyperref[prompt:3]{Prompt 3} employs an LLM-as-a-judge methodology to evaluate the models' responses to the questions. 
Prompts used for processing policies from different countries were designed to correspond to the respective national language. Only the English prompts are presented here. Prompts were translated using Google Translate to ensure consistency between the two languages. All translated content was subsequently reviewed and tested by human annotators to avoid potential semantic inconsistencies.

\clearpage
\label{sec:prompt}

\begin{figure*}[t]
\centering
\begin{tcolorbox}[prompt, title=Prompt 1: Level-1 cloze-style Generation, colback=white, width=\textwidth]
\label{prompt:1}
You are a policy expert and your task is to generate questions based on the given policy text.\\
- Strictly follow the given material to generate questions, do not fabricate content.\\
- Generate 5-10 fill-in-the-blank questions and 3-8 true or false questions based on the length of the policy.\\
- Answer to the questions should be clear and precise, avoid ambiguous answers.\\
- Answer to the questions should preferably be a single word or phrase.\\
- If the answer is a false judgment question, Avoid altering, adding, or deleting the original text when the answer is not an incorrect judgment question.\\
- Don't generate questions related to non-key information such as file numbers.\\
- Each question should start with "According toXXX", please provide the full name of the policy.\\
- Questions should not be limited to a single aspect (such as time), and should be diversified.\\

policy title: \{policy\}\\
policy content: \{policy\_content\}
\end{tcolorbox}
\end{figure*}

\begin{figure*}[t]
\centering
\begin{tcolorbox}[prompt, title=Prompt 2: Distractor generation, colback=white,width=\textwidth]
\label{prompt:2}
You are an ai assistant tasked with answering policy-related questions.\\

- Answer the questions based on your knowledge.\\
- Please note that some incorrect answers are provided below. You must not make the same mistakes, \\
- Your answer needs to be semantically distinct from the given incorrect answer.\\
- Don't say you can't see the image, just answer based on your knowledge.\\
- Don't generate overly lengthy answers, keep them concise and to the point. \\ 
- The answer you generate needs to be factually different from the given incorrect answer.\\
- Try to use straightforward words instead of being too abstract or vague.\\

question:\{question\}\\
wrong answers:\{wrong\_answer\}
\end{tcolorbox}
\end{figure*}

\begin{figure*}[t]
\centering
\begin{tcolorbox}[prompt, title=Prompt 3: LLM-as-a-Judge, colback=white, width=\textwidth]
\label{prompt:3}
    You are an expert evaluator. Your task is to score the following open-ended answer based on a reference answer and scoring criteria. Follow these rules carefully:\\
    1. For calculation or factual questions where the result must be precise (e.g., math, unit conversion, logical problems), if the final answer is incorrect, the score should be 0, regardless of the explanation.\\
    2. For general questions (e.g., reasoning, explanation, analysis), the reference answer includes multiple key points. \\
        - Compare the given answer with the reference key points.\\
        - For each matched key point, assign partial credit proportionally.\\
        - If the answer includes correct but unlisted points (beyond the reference answer), you may award partial credit with explanation.\\
    3. Provide a score from 0 to 5. Generally:\\
        - 5 = Completely correct and well explained\\
        - 4 = Mostly correct, with minor issues\\
        - 3 = Partially correct, some key points missing or wrong\\
        - 2 = Mostly incorrect but with small redeeming aspects\\
        - 1 = Barely relevant or correct\\
        - 0 = Completely wrong or irrelevant\\
    4. In your reasoning, clearly list:\\
        - Which points in the reference answer are matched\\
        - Any extra correct points beyond the reference\\
        - Justify any deductions\\
    5. Be strict but fair. Do not be lenient.\\
    ---\\
    Question: \{question\}  \\
    Reference Answer:\\
    \{reference\_answer\_with\_point\_marks\}\\
    User Answer:
    \{user\_answer\}\\
    ---\\
    Now output:\\
    Score: X  \\
    Reasoning: ...\\
\end{tcolorbox}
\end{figure*}

\end{document}